\documentclass{article} \usepackage{iclr2026_conference,times}

\usepackage{amsmath,amsfonts,bm}

\usepackage{url}
\usepackage{wrapfig}
\usepackage{dblfloatfix}
\usepackage[linesnumbered,ruled,vlined]{algorithm2e}
\usepackage{array}
\usepackage{balance}
\usepackage{booktabs}
\usepackage{caption}
\usepackage{enumitem}
\usepackage{flushend}
\usepackage{float}
\usepackage{graphicx}
\usepackage{hyperref}
\usepackage{makecell}
\usepackage{multirow}
\usepackage{mathtools}
\usepackage{pifont}
\usepackage{subcaption}
\usepackage{tabularx}
\usepackage{tabularray}
\usepackage{threeparttable}
\usepackage{xcolor}
\usepackage{placeins}

\newcommand{\name}{CacheRepair}
\newcommand{\titletext}{Learning to Repair Cross-Chunk Context in RAG for KV Cache Fusion}
\title{\name{}: \titletext}

\iclrfinalcopy
\author{
Genglin Wang\textsuperscript{1}, Wangsong Yin\textsuperscript{2},
Yeerzhati Abudunuer\textsuperscript{1} \\
\textbf{Haoxuan Xu\textsuperscript{3}, Guoliang Xing\textsuperscript{1},
Zhenyu Yan\textsuperscript{1}} \\
\textsuperscript{1}The Chinese University of Hong Kong \\
\textsuperscript{2}Peking University \quad
\textsuperscript{3}The Hong Kong University of Science and Technology
}
\hypersetup{
  hidelinks,
  pdftitle={CacheRepair: Learning to Repair Cross-Chunk Context in RAG for KV Cache Fusion},
  pdfauthor={Genglin Wang, Wangsong Yin, Yeerzhati Abudunuer, Haoxuan Xu, Guoliang Xing, Zhenyu Yan}
}

\begin{document}
\raggedbottom

\maketitle
\lhead{Preprint}

\begin{abstract}
Multi-document retrieval-augmented generation (RAG) requires a
language model to process multiple retrieved text chunks before answering a
question. Precomputing each chunk's KV cache independently and concatenating
the caches when the chunks are retrieved can accelerate this step. However,
the assembled cache lacks cross-chunk attention information, reducing answer
quality. Selective recomputation methods recover the missing cross-chunk
context by rerunning the target LLM on selected tokens, incurring substantial
online computation. We introduce \name{}, a lightweight network that learns
the difference between independently computed KV caches and those produced
by processing the chunks together. The network combines compressed KV
features with token embeddings and uses attention that is bidirectional
within each chunk and flows from earlier to later chunks. Each repair block receives the compressed cache features, and the predicted residual is added to every document token's cache. Each repair network is trained for
a specific frozen target LLM on a generic retrieval corpus and reused
across downstream datasets. Our analysis shows that repair reduces KV errors
both near chunk boundaries and throughout chunk interiors. Evaluation across
three target LLMs and four downstream datasets places
\name{} on the measured answer-quality--latency Pareto frontier in eleven
of twelve model--dataset combinations. Reported time to first token (TTFT) includes online cache transfer
and repair. Across all twelve combinations, the largest repairers achieve
1.69--4.61$\times$ speedups in median TTFT over full prefill and improve
mean F1 by 2.1--26.1 percentage points over direct cache reuse.
\end{abstract}
\vspace{-2pt}
{\centering\setlength{\parskip}{0pt}
\textbf{Links:}\quad
\href{https://github.com/genglinWang/CacheRepair-code}{\textcolor{magenta}{Code} (GitHub)}
\quad\textbar\quad
\href{https://huggingface.co/gwang3456/cacherepair}{\textcolor{magenta}{Models} (Hugging Face)}
\par}

\section{Introduction}

\begin{figure}[!b]
    \centering
    \vspace{-10pt}
    \includegraphics[width=1.0\linewidth]{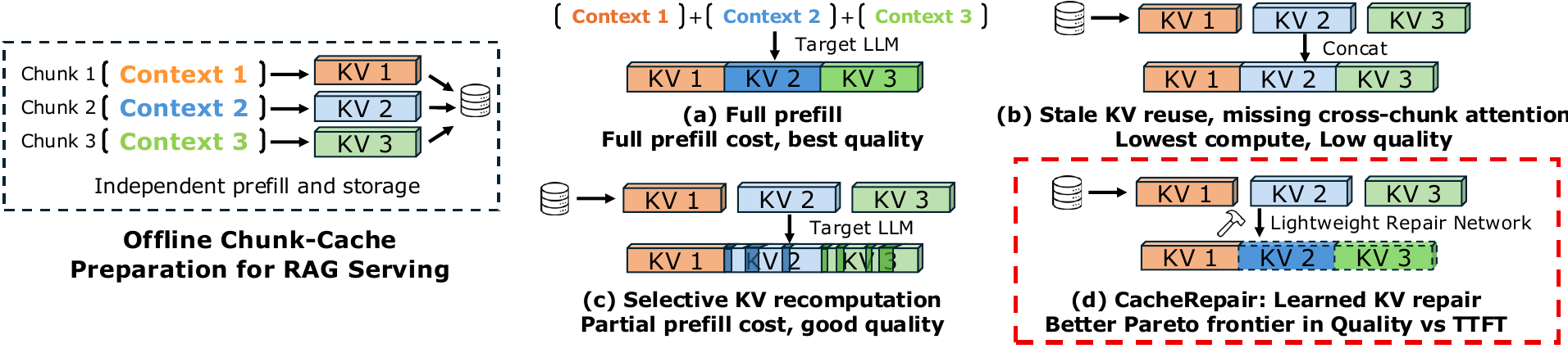}
    \caption{\name{}'s paradigm: a lightweight KV cache repair network.}
    \label{fig:teaser}
    \vspace{-14pt}
\end{figure}

Retrieval-augmented generation (RAG) is widely used for
knowledge-intensive question answering, grounding language-model responses
in retrieved evidence \citep{lewis2021rag,izacard2021fid}.
In multi-document RAG, a large language model (LLM) processes several retrieved
chunks during \emph{prefill} before answering, adding to time to first token
(TTFT). To avoid repeating this computation when chunks recur across requests,
their KV caches can be computed independently in advance, stored, and
concatenated when retrieved
\citep{yao2025cacheblend,hu2025epic,lu2024turborag}.
This enables reuse across different chunk combinations and orders.
However, each chunk is precomputed without access to the other chunks in a
future request, leaving cross-chunk context missing from the concatenated cache.
We call the concatenated cache the \emph{stale KV cache} (Figure~\ref{fig:teaser}(b)), and
the jointly computed reference the \emph{joint KV cache}
(Figure~\ref{fig:teaser}(a)). In this paper, we address the challenge of
recovering the missing cross-chunk context with low online computation.

Existing KV-fusion methods address this missing cross-chunk
context in several ways. As we show in Figure~\ref{fig:teaser}(c), KV recomputation recomputes selective document tokens with the target LLM, using an online token budget
to restore contextual information
\citep{yao2025cacheblend,hu2025epic,teng2026infoflow}.
Other methods reshape attention, adapt the target LLM, or learn reusable
cache components during offline preparation
\citep{yang2025ape,lu2024turborag,ma2025blockattention,yang2025kvlink,chen2026kvpacket}.
We investigate a central question: can a lightweight
external network recover cross-chunk context directly in the stale KV,
while preserving the frozen LLM and reusable caches?

Our empirical study of Qwen2.5-3B reveals structure in the
difference between stale and joint KV. Errors form both
\emph{boundary-local hotspots} near chunk beginnings and
\emph{layer-persistent bands} extending across token positions in particular
layers (Section~\ref{sec:observations}). Thus, the discrepancy extends into
chunk interiors as well as boundaries. Selective recomputation updates a
subset of these positions; the broader error structure motivates repair
across all document tokens. At the same time, stale KV already contains
the computation performed within each chunk. These observations motivate
using the existing cache as the starting point and learning the residual
needed to approximate joint KV.

\begin{figure}[!t]
\centering
\includegraphics[width=1.0\linewidth]{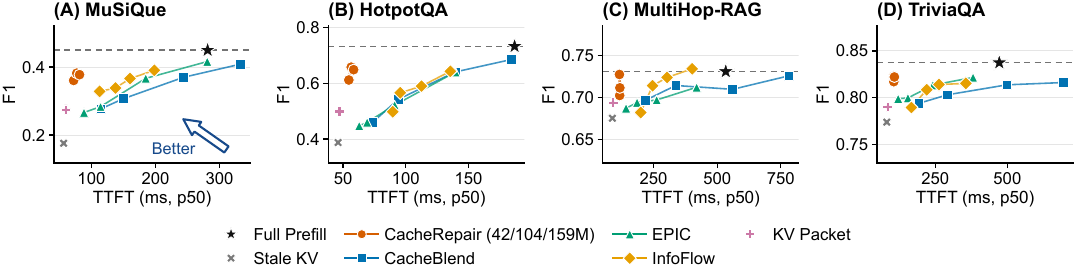}
\caption{Answer F1 versus p50 TTFT for Qwen2.5-14B on four downstream datasets.}\label{fig:qwen14-four-dataset-pareto}
\vspace{-8pt}
\end{figure}

We introduce \name{}, the first lightweight network that predicts
the residual between joint and stale KV caches
for every document token (Figure~\ref{fig:teaser}(d)).
Our key design replaces selective target-LLM recomputation
with direct residual prediction by an external network, reducing online
work while keeping the target LLM frozen.
The \emph{StaleEncoder} compresses KV features from all target-LLM layers
and combines them with projected frozen token embeddings, providing cache
content and token identity. \emph{Block-causal attention}
allows bidirectional interaction within each chunk and passes information
from earlier to later chunks. Each repair block receives a learned projection
of the original cache features, maintaining cache conditioning throughout
repair. An output skip connection adds the predicted residual to stale KV. For keys, we remove
local rotary positional encoding (RoPE)~\citep{rope} before repair and apply global RoPE
afterward, separating position alignment from learned context repair;
values retain their original coordinates.

For each target LLM, we train repairers on a generic retrieval
corpus and evaluate each fixed checkpoint on four downstream datasets:
MuSiQue, HotpotQA, MultiHop-RAG, and TriviaQA. The same repairer serves
all four downstream workloads. Our evaluation spans Qwen2.5-3B,
Llama-3.1-8B, and Qwen2.5-14B. CacheRepair contributes at least one
nondominated operating point in eleven of the twelve target--workload pairs
(Figures~\ref{fig:qwen14-four-dataset-pareto},
\ref{fig:qwen3-four-dataset-pareto}, and
\ref{fig:llama8-four-dataset-pareto}). For example, Figure~\ref{fig:qwen14-four-dataset-pareto}
shows that three repairer sizes (42M, 104M, and 159M) for Qwen2.5-14B
extend the quality--latency Pareto frontier of the evaluated baselines
across all four downstream datasets. Across all twelve model--dataset
combinations, the largest repairers deliver 1.69--4.61$\times$ speedups in
p50 TTFT over Full Prefill and improve mean F1 by 2.1--26.1 percentage points
over direct cache reuse. These latency measurements include online cache
transfer, repair, and target-LLM processing through first-token generation.
These results demonstrate that a repairer trained on generic
retrieval data improves reused-cache quality across downstream tasks.

\begin{itemize}[leftmargin=*,nosep]
    \item \textbf{Characterizing cross-chunk KV error.} We identify structured boundary-local
    and layer-persistent stale-KV error patterns that motivate repair across document positions.
    \item \textbf{Learning all-token KV residual repair.} We develop a
    learned network that combines stale-cache features,
    token identity, and chunk-structured attention to repair KV for a frozen LLM.
    \item \textbf{Improving quality at low latency.}
    Across twelve model--dataset combinations, the largest repairers achieve
    1.69--4.61$\times$ TTFT speedups over Full Prefill and gain 2.1--26.1 F1
    percentage points over direct cache reuse, reusing each checkpoint across
    downstream tasks.
\end{itemize}

\section{Related Work and Problem Setting}
\subsection{Related Work}

\noindent\textbf{Multi-Document RAG.}
Retrieval-augmented generation combines retrieved evidence
with language generation for knowledge-intensive answering
\citep{lewis2021rag,izacard2021fid}. Retrieving multiple documents supplies
broader evidence but also lengthens the input that the model must
process before generating its first token. This additional prefill
computation can substantially increase TTFT \citep{lu2024turborag}.
Documents that recur across requests therefore offer an opportunity
to reduce this cost by reusing precomputed KV caches.

\noindent\textbf{KV Cache Reuse.}
Reusing previously computed KV states reduces repeated prompt processing.
PromptCache organizes reusable prompt modules
\citep{gim2024promptcachemodularattention}, while SGLang shares cached
prefixes through a radix-tree structure \citep{heng2024sglang}.
For multi-document RAG, independently precomputing each chunk's cache enables
reuse across requests that retrieve different chunk combinations and orders
\citep{lu2024turborag,yao2025cacheblend}. This flexibility introduces a
context mismatch: independently cached chunks omit attention to preceding
chunks in the assembled request. KV cache fusion addresses how to recover
answer quality when these caches are used together.

\noindent\textbf{KV Cache Fusion.}
One approach recovers cross-chunk context through selective target-LLM
recomputation. CacheBlend selects tokens using KV deviation
\citep{yao2025cacheblend}; EPIC focuses computation near chunk boundaries
\citep{hu2025epic}; and InfoFlow uses an information-flow signal
\citep{teng2026infoflow}. These methods control online target-LLM computation
through the selected token budget. Related approaches combine partial recomputation
with cache management or auxiliary-model token selection
\citep{agarwal2025cachecraft,yang2025cacheclip}. Other methods reshape
attention \citep{yang2025ape} or adapt the target LLM for independently
encoded chunks and inter-document links
\citep{lu2024turborag,ma2025blockattention,yang2025kvlink}.
KV Packet learns reusable soft-token headers and trailers during offline
cache construction \citep{chen2026kvpacket}; Cartridges learns corpus-specific
KV representations through offline self-study \citep{eyuboglu2025cartridges}.
\name{} uses a target-specific external network to reconstruct the frozen
target LLM's joint KV states by predicting residual corrections for every
document token in the stale cache. The repaired KV entries provide
cross-chunk context for query processing and answer generation.

\subsection{Problem Setting and Objectives}
\label{sec:problem-setting}

\noindent\textbf{Independent Reuse and Stale Cache.}
A request $r$ contains an ordered document sequence
$D_r=[C_{\pi_r(1)},\ldots,C_{\pi_r(m)}]$ followed by a query.  Jointly
prefilling $D_r$ with a frozen target LLM $M$ produces the request-specific
joint-prefill KV cache $KV_{\mathrm{joint},r}$.  Independent reuse instead
prefills each chunk $C_i$ once and
stores its independently prefilled KV cache $KV_{\mathrm{ind}}(C_i)$.  At
serving time, the selected independently prefilled KV caches are
aligned to global request positions by re-rotating K and concatenated into the stale KV cache
$KV_{\mathrm{stale},r}$.
The stale KV cache preserves each chunk's independently
computed KV cache but lacks the cross-chunk attention information in joint
prefill; position alignment alone cannot recover that information, so
$KV_{\mathrm{stale},r}\neq KV_{\mathrm{joint},r}$.

\noindent\textbf{Learned KV-Cache Repair Target.}
We study a lightweight learned operator $\mathcal{R}_\theta$ that receives the
stale KV cache together with the document tokens and chunk layout, and
produces
\begin{equation*}
    \widehat{KV}_r
    = \mathcal{R}_\theta(KV_{\mathrm{stale},r},D_r)
    \approx KV_{\mathrm{joint},r}.
\end{equation*}
The joint-prefill KV cache serves as an offline learning and evaluation target;
the target LLM remains frozen, and repair depends on the ordered documents,
not the query text.

\noindent\textbf{System Objectives and Constraints.}
Repair should approach joint-prefill quality by correcting cached KV directly,
while keeping the target LLM frozen and retaining offline chunk reuse.
We next examine the structure of cross-chunk KV error to guide the design of learned repair.

\section{Observation}
\label{sec:observations}

We examine where independently computed caches differ from
joint prefill, asking which document positions and target-LLM layers need
context repair.

\begin{figure*}[t]
  \centering
  \includegraphics[width=1.0\textwidth]{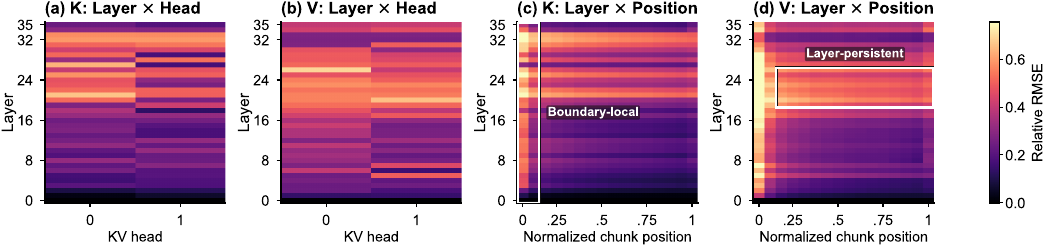}
  \caption{Structured stale-KV error in Qwen2.5-3B on the first 256 MuSiQue requests.
  (a--b) Relative RMSE by layer and KV head. (c--d) Relative RMSE by layer and
  within-chunk position; outlines highlight boundary-local and layer-persistent
  patterns.}
  \label{fig:motivation-observations}
  \vspace{-10pt}
\end{figure*}

Using frozen Qwen2.5-3B-Instruct (36 layers and two KV heads),
we compare stale and joint KV on the first 256 MuSiQue
requests \citep{trivedi-etal-2022-musique}. The requests contain
579,867 document tokens; the heatmaps cover the 545,276 tokens in chunks
after the first, where preceding-document context becomes available during
joint prefill.
For each cell, relative RMSE is the square root of the ratio of summed
squared KV error to summed squared joint-KV magnitude, pooling the corresponding
entries across examples. Figure~\ref{fig:motivation-observations}(a--b)
groups entries by layer and KV head; panels (c--d) use 16 normalized
within-chunk position bins, pooling both KV heads.

\textbf{Cross-chunk KV error has boundary-local and
layer-persistent structure.} The layer--head maps show larger relative
errors in several middle and upper layers. Resolving these errors by token
position reveals two patterns. In both K and V, elevated error near chunk beginnings
forms a vertical hotspot, and high-error bands span boundary and interior
positions in particular layers. Thus, recovering cross-chunk context
involves positions throughout the document, with error magnitude varying
across layers and within-chunk locations. Appendix~\ref{app:observation-generality}
shows these patterns across all three target LLMs.

This structure motivates learning a correction for every
document token, including interior positions in layer-persistent error bands.
Making all-token repair practical also requires keeping its
online cost low. Selective recomputation limits this cost by rerunning the
target LLM on only a subset of document tokens~\citep{yao2025cacheblend,hu2025epic}.
For an $N$-token document cache and a selected fraction $r$, attention
between the $rN$ selected tokens and the cache costs $O(rN^2)$ at fixed
target-LLM dimensions. Our repair network processes all tokens using fewer
layers and smaller hidden dimensions. Its attention is also quadratic in
$N$, so the cost comparison depends on network dimensions, token budgets,
kernels, and cache movement. Section~\ref{sec:repair-mechanism} and
Appendix~\ref{app:scaling-amortization} quantify these costs across lengths.
The stale cache already supplies chunk-local representations, allowing the
repairer to focus its capacity on predicting the difference to joint KV.
We therefore combine compressed stale-cache features, token information,
and chunk-structured attention to provide broad repair coverage at low online
cost (Section~\ref{sec:method}).

\section{Method}
\label{sec:method}

\begin{figure*}[t]
    \centering
    \includegraphics[width=1.0\textwidth]{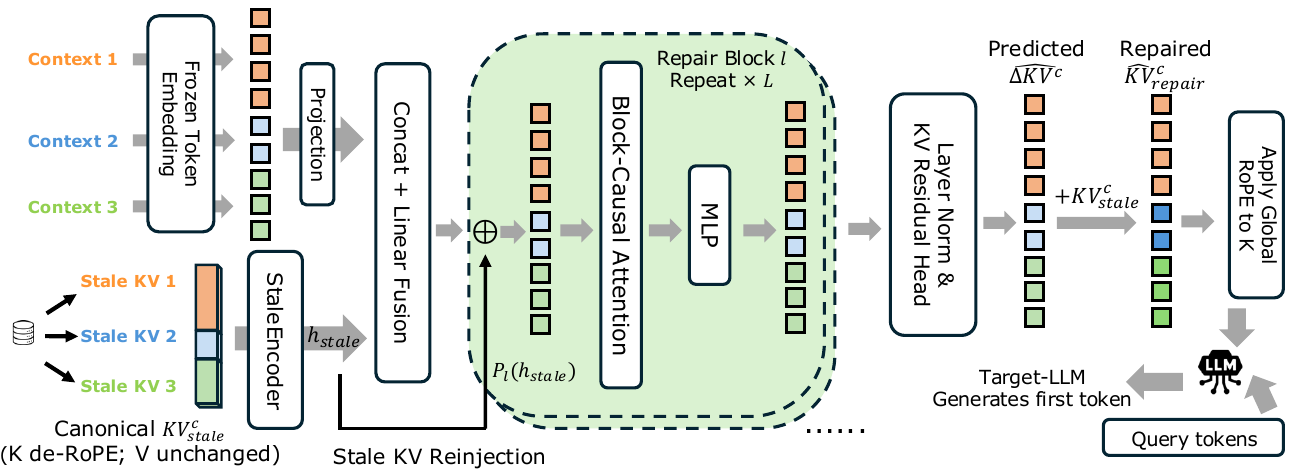}
    \caption{\textbf{CacheRepair inference pipeline.}
    Compressed stale-KV features and token embeddings condition the repair
    blocks. Each block receives stale-feature reinjection; the predicted
    residual is added to the full stale KV before global RoPE is applied to K.}
    \label{fig:cacherepair-architecture}
\end{figure*}

As shown in Figure~\ref{fig:cacherepair-architecture},
\name{} takes the assembled stale KV and the corresponding document tokens
as inputs. It compresses the cache and combines it with token embeddings
from the frozen target LLM. Repair blocks exchange information across chunks, with the
original cache features supplied to each block. The output head predicts
KV residuals for every document token, which are added to the stale cache.
After global positional
encoding is applied to keys, the frozen target LLM uses the repaired cache
to process the query and generate an answer.

\subsection{Inference}

\noindent\textbf{Chunk-cache preparation.}
Each chunk is independently prefilled and cached in advance. To support its
placement in different retrieved sequences, we store K in \emph{canonical}
coordinates, obtained by removing local rotary positional encoding (RoPE).
Let $R_M(p)$ denote the target LLM's native RoPE transformation at position
$p$. For a token $t$ at local position $q_t$, the stored entries are
\begin{equation*}
    K_{\mathrm{ind},t}^c=R_M(q_t)^{-1}K_{\mathrm{ind},t},
    \qquad V_{\mathrm{ind},t}^c=V_{\mathrm{ind},t}.
\end{equation*}
\noindent\textbf{Coordinate convention.}
Throughout the method, $KV^c$ abbreviates the pair $(K^c,V)$: the coordinate
transformation applies to K, while V is stored and predicted in its native
coordinates. Any $V^c$ notation denotes this same untransformed V.

At serving time, retrieved caches are
concatenated in document order to form $KV^c_{\mathrm{stale}}$. The target
LLM's rotary frequencies and scaling parameters determine $R_M$.

\noindent\textbf{StaleEncoder and token fusion.}
The StaleEncoder compresses all layers of a token's KV into a feature vector
$s_t$ of width $d_b$. For a target with $L_M$ layers, $H_{\mathrm{KV}}$ KV
heads, and head dimension $d_h$, each token has $S=2L_MH_{\mathrm{KV}}$
segments, one per layer, head, and K/V component. Each segment has its own
learned $d_h\!\rightarrow\!d_{\mathrm{seg}}$ projection. The encoder concatenates these outputs
and maps them to $d_b$. Its input is scaled by fixed per-coordinate RMS
statistics $\sigma_{\mathrm{stale}}$, defined in Section~\ref{sec:repair-training}.
In parallel, a trainable projection $P_{\mathrm{tok}}$ maps the document
token's frozen target embedding $E_M(x_t)$ to width $d_b$.
A linear fusion layer $F:\mathbb{R}^{2d_b}\!\rightarrow\!\mathbb{R}^{d_b}$ combines cache content and token identity:
\begin{equation*}
    s_t=\operatorname{StaleEncoder}
      (KV^c_{\mathrm{stale},t}/\sigma_{\mathrm{stale}}),
    \qquad
    h_t^0=F\bigl([s_t;P_{\mathrm{tok}}E_M(x_t)]\bigr).
\end{equation*}
Here $[\,;\,]$ denotes feature concatenation and division is elementwise.
The sequence $h^0$ is the input to the repair backbone.

\noindent\textbf{Repair Blocks and stale-feature reinjection.}
For a sequence of $N$ document tokens, the backbone applies $B$ Repair
Blocks of width $d_b$, each combining attention and an MLP. Let $c(t)$ be the index of token $t$'s chunk in request
order. The block-causal mask permits token $i$ to attend to token $j$ when
\begin{equation*}
    A_{ij}=\mathbf{1}[c(j)\leq c(i)].
\end{equation*}
Every retrieved chunk is fully available at repair time, enabling
bidirectional interaction within it. Across chunks, information flows from
earlier to later chunks, matching the direction of the missing context.
Each block applies the repair backbone's own RoPE to attention Q and K at
absolute document positions $p=(p_1,\ldots,p_N)$.

To keep the cached representations available throughout this interaction,
each block receives a learned projection of the original encoded features:
\begin{equation*}
    h^b=\operatorname{RepairBlock}_b
       \bigl(h^{b-1}+P_b(s);A,p\bigr),\qquad b=1,\ldots,B.
\end{equation*}
The encoded sequence $s=(s_1,\ldots,s_N)$ is computed once; each block has its own
projection $P_b$. This reinjection conditions the evolving hidden states on
stale-cache content. The full stale KV is also preserved for the output
addition described next.

\noindent\textbf{KV residual output and global RoPE.}
Layer normalization and a linear KV Residual Head map $h_t^B$ to a normalized
residual $z_t$ with $D_{\mathrm{KV}}=S d_h$ coordinates.
Appendix~\ref{app:residual-capacity} analyzes residual structure, output-space
capacity, and parameter allocation. Using the fixed
residual scale $\sigma_\Delta$ (Section~\ref{sec:repair-training}), we recover the residual in KV units and add
it to the complete stale cache:
\begin{equation*}
    z_t=\operatorname{KVResidualHead}(\operatorname{LN}(h_t^B)),
    \qquad
    \widehat{KV}_{\mathrm{repair},t}^c
       =KV_{\mathrm{stale},t}^c+\sigma_\Delta\odot z_t.
\end{equation*}
This output skip connection retains the original KV entries while the
network supplies their learned correction. Both terms use the canonical
coordinate convention. We then place K at its global request position:
\begin{equation*}
    \widehat K_{\mathrm{repair},t}
       =R_M(p_t)\widehat K_{\mathrm{repair},t}^c,
    \qquad
    \widehat V_{\mathrm{repair},t}
       =\widehat V_{\mathrm{repair},t}^c.
\end{equation*}
The repairer processes every document token, including the first chunk.
For that chunk, the independent and joint contexts coincide, giving a
near-zero residual target; Appendix~\ref{app:repair-mechanism} reports its
measured error before and after repair. The frozen
target LLM consumes the repaired cache when processing the query and
generating the answer; the serving integration is described in
Section~\ref{sec:experiments}.

\subsection{Training}
\label{sec:repair-training}

\noindent\textbf{Paired cache supervision.}
For each ordered document sequence in the training corpus, the frozen target
LLM produces both independently prefilled chunk caches and a joint-prefill
cache for the identical tokens. We remove local RoPE from independently
computed K and global RoPE from jointly computed K, placing both in
canonical coordinates. Their difference provides the supervision:
\begin{equation*}
    \Delta KV_t^{c\star}
       =KV_{\mathrm{joint},t}^c-KV_{\mathrm{stale},t}^c.
\end{equation*}
This target teaches the repairer to reconstruct the cross-chunk change in
cached representations from document inputs.

\noindent\textbf{Normalization and objective.}
We compute $\sigma_{\mathrm{stale}}$ and $\sigma_\Delta$ by taking RMS values
across training-corpus document tokens separately for each target layer,
KV head, K/V component, and head coordinate. These scales normalize the
input cache and target residual, respectively, and are stored with the
checkpoint for serving. For an example with $N$ document tokens, training
minimizes mean squared error in normalized coordinates:
\begin{equation*}
    \mathcal L_{\mathrm{repair}}
       =\frac{1}{N D_{\mathrm{KV}}}\sum_{t=1}^{N}
         \left\|z_t-\Delta KV_t^{c\star}/\sigma_\Delta\right\|_2^2.
\end{equation*}
Every document token and normalized KV coordinate receives equal weight.
We optimize the StaleEncoder, token projection and fusion layer, Repair
Blocks, reinjection projections, and output head. The resulting checkpoint
pairs a target-specific repair network with fixed normalization statistics
and is reused across downstream datasets.

\section{Experiments}
\label{sec:experiments}

\subsection{Experimental setup}
\label{sec:setup}

\noindent\textbf{Models and training.}
We evaluate Qwen2.5-3B-Instruct and Qwen2.5-14B-Instruct~\citep{qwen2025technicalreport},
and Llama-3.1-8B-Instruct~\citep{grattafiori2024llama3},
with three repairer capacities per target
(Table~\ref{tab:repair-scales}). We evaluate each repairer after six epochs on
a generic retrieval corpus drawn from six sources
(Appendix~\ref{app:generic-corpus}). We use its epoch-six checkpoint and
normalization statistics across all four downstream datasets. The target
LLM remains frozen throughout training and evaluation.

\textbf{Datasets and baselines.}
We evaluate 500 requests from each of four downstream datasets:
MuSiQue~\citep{trivedi-etal-2022-musique},
HotpotQA~\citep{yang-etal-2018-hotpotqa},
MultiHop-RAG~\citep{tang2024multihoprag}, and
TriviaQA~\citep{joshi-etal-2017-triviaqa}.
We screen questions and context against the training corpus to ensure a
fair evaluation (Appendix~\ref{app:cohort-construction}). All targets use
the same questions; all methods use identical inputs for a given target.
Each request contains at most 4,096 document tokens and 20 chunks.
We compare CacheRepair with Full Prefill, direct Stale KV reuse,
CacheBlend~\citep{yao2025cacheblend}, EPIC~\citep{hu2025epic},
InfoFlow~\citep{teng2026infoflow}, and KV Packet~\citep{chen2026kvpacket}.
KV Packet uses a shared subset of 512 examples from the repair-training corpus
(Appendix~\ref{app:learned-baseline-compatibility}).
Appendix~\ref{app:evaluation-details} gives the input preparation and
configuration grids.

\textbf{Execution and metrics.}
The main evaluation runs one request at a time on one NVIDIA A800-SXM4-80GB, recording
F1, EM, and TTFT from the same generation. TTFT includes cache transfer, positional alignment,
repair or token selection and recomputation, query processing, and
first-token generation; independent cache construction and host preparation
are completed beforehand. We report mean F1 and p50 TTFT; p90 and p99
latencies are included in the accompanying data. We bootstrap requests to
estimate uncertainty and use paired draws for method comparisons.
Appendix~\ref{app:evaluation-details} specifies the runtime, timing, and
statistical procedures.

\subsection{Results}
\label{sec:results}

The Pareto plots show all tested configurations on the
hardware and runtime specified above. Frontier membership uses mean F1
and p50 TTFT point estimates. Lines connect
settings in the order listed in Appendix~\ref{app:evaluation-details};
dashed lines mark Full Prefill F1. Panels use separate axis ranges.

\begin{figure}[t]
\centering
\includegraphics[width=\linewidth]{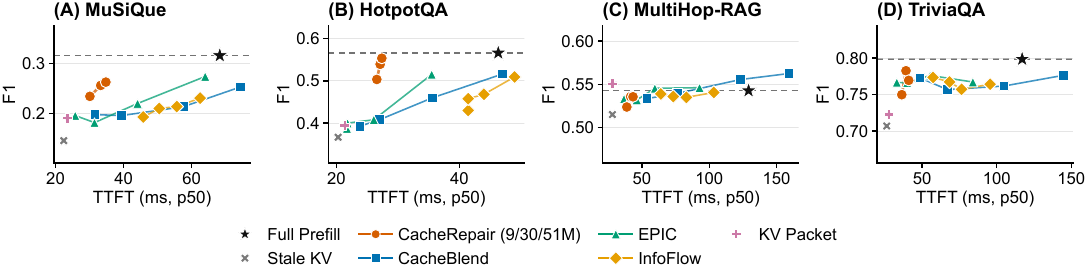}
\caption{Answer F1 versus p50 TTFT for Qwen2.5-3B on four downstream datasets.}
\label{fig:qwen3-four-dataset-pareto}
\end{figure}

\noindent\textbf{How does CacheRepair compare with the baselines?}
CacheRepair contributes quality--TTFT frontier points in
11 of the 12 model--dataset settings among the evaluated configurations.
These comparisons include CacheBlend,
EPIC, InfoFlow, and KV Packet. Figure~\ref{fig:qwen3-four-dataset-pareto}
shows frontier points on three Qwen2.5-3B datasets. On MultiHop-RAG,
KV Packet has lower TTFT and higher mean F1 in the main 500-request batch;
Appendix~\ref{app:additional-requests} reports an independent batch and
paired quality intervals. These results show that CacheRepair advances
the measured quality--latency frontier across target LLMs and downstream
datasets.

\noindent\textbf{How much quality is retained at half the Full Prefill latency?}
We set a budget of $0.5T_{\mathrm{full}}$ and select,
for each method, its largest tested capacity or recomputation setting
whose p50 TTFT meets the budget. Selection uses latency alone.
Table~\ref{tab:budget-comparisons} summarizes all twelve settings.
CacheRepair's mean F1 exceeds every eligible baseline in nine;
simultaneous paired intervals remain positive in four. The remaining
three settings use Qwen2.5-3B: no repairer meets the HotpotQA budget,
and the highest eligible means on MultiHop-RAG and TriviaQA belong to
KV Packet and EPIC, respectively. Appendix~\ref{app:cross-model-budget}
provides the per-method scores.

\begin{figure}[t]
\centering
\includegraphics[width=\linewidth]{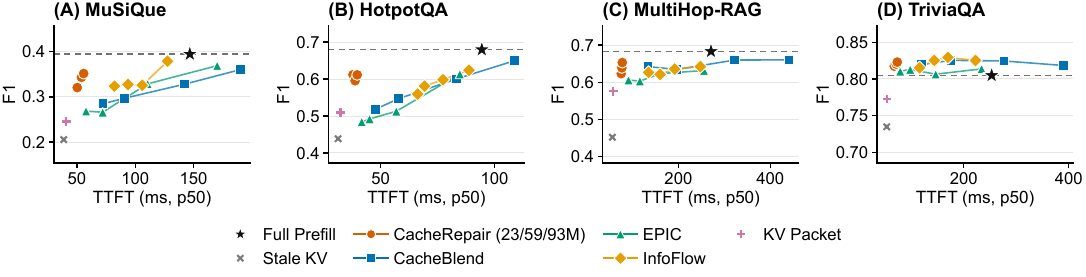}
\caption{Answer F1 versus p50 TTFT for Llama-3.1-8B on four downstream datasets.}
\label{fig:llama8-four-dataset-pareto}
\end{figure}

\noindent\textbf{How do results vary across target LLMs?}
CacheRepair contributes frontier points on all four
datasets for both Llama-3.1-8B and Qwen2.5-14B
(Figures~\ref{fig:llama8-four-dataset-pareto} and
\ref{fig:qwen14-four-dataset-pareto}). On Qwen2.5-14B/MultiHop-RAG,
the 159M repairer achieves 72.71\% F1 at 115.74\,ms, compared with
72.55\% for CacheBlend 0.6, giving a $6.76\times$ TTFT speedup.
The paired F1 difference is $+0.16$ points with a 95\% interval
[$-1.95$, $2.26$], with closely matched mean answer F1.

Across all twelve settings, the largest repairers are
1.30--6.76$\times$ faster than the highest tested recomputation presets
(CacheBlend 0.6, EPIC64, and InfoFlow 0.25). Their F1 differences range
from $-3.86$ to $+4.43$ points; Appendix~\ref{app:cross-model-budget}
reports the comparison scope and uncertainty. Relative to the reference
methods, these repairers improve Stale KV F1 by 2.1--26.1 points and
provide 1.69--4.61$\times$ speedups over Full Prefill.
Table~\ref{tab:quality-tradeoffs} gives the corresponding Full Prefill
quality differences for every setting.

\noindent\textbf{What contributes to CacheRepair's TTFT?}
Table~\ref{tab:repair-latency-components} reports measurements
from the largest repairer for each target on MultiHop-RAG.
Median repair execution takes 12.55/\allowbreak 18.14/\allowbreak 26.52\,ms for
Qwen2.5-3B/Llama-3.1-8B/Qwen2.5-14B, while host-to-device transfer takes
6.30/\allowbreak 20.78/\allowbreak 33.08\,ms. Cache transfer takes longer than repair on
Llama-3.1-8B and Qwen2.5-14B, reflecting the cost of moving larger KV tensors.
From 3B to 14B, repair time grows only $2.11\times$, while Full Prefill
TTFT grows $4.13\times$. This slower growth favors larger target LLMs:
the speedup over Full Prefill rises from $2.98\times$ to $4.61\times$
on this workload.

\begin{table}[t]
\centering
\caption{Median online latency components (ms) for the
largest repairers on MultiHop-RAG. TTFT also includes query processing,
first-token generation, and runtime overhead. Component medians do not sum
to median TTFT.}
\label{tab:repair-latency-components}
\small
\setlength{\tabcolsep}{4pt}
\begin{tabular}{lrrrrrr}
\toprule
Target LLM & H2D & Mask & Repair & RoPE & KV write & TTFT \\
\midrule
Qwen2.5-3B & 6.30 & 1.25 & 12.55 & 2.17 & 1.29 & 43.41 \\
Llama-3.1-8B & 20.78 & 1.37 & 18.14 & 6.56 & 3.63 & 77.85 \\
Qwen2.5-14B & 33.08 & 1.44 & 26.52 & 10.01 & 5.60 & 115.74 \\
\bottomrule
\end{tabular}
\end{table}

\noindent\textbf{What do the design choices contribute?}
Table~\ref{tab:controlled-ablation} compares six variants
after two epochs (25,000 updates) on the same training data and MuSiQue
requests. A0, the full design, reaches 0.2462 F1. Block-diagonal attention
reaches 0.0133, supporting the importance of cross-chunk information exchange.
Removing the residual output path while retaining the same normalization
reaches 0.0049. With joint-RMS normalization, direct prediction reaches
0.0118 after the same two epochs, giving A0 a 23.44-point mean F1 advantage
(Appendix~\ref{app:attention-topology}). The token-causal, bidirectional,
and entrance-only variants differ from A0 by at most 0.62 F1
percentage points in Table~\ref{tab:controlled-ablation}.
Block-causal attention enables cross-chunk information exchange while
preserving document order. It achieves the highest mean F1 among the
evaluated attention variants. Appendix~\ref{app:attention-topology} reports regional KV errors and training details.

\begin{table}[t]
\centering
\caption{Six repairer variants trained for two epochs and evaluated on the same 500 MuSiQue requests with Qwen2.5-3B. $\Delta$F1 is computed from the reported mean F1 values relative to A0, in percentage points.}
\label{tab:controlled-ablation}
\small
\setlength{\tabcolsep}{4pt}
\begin{tabular}{lrrrrr}
\toprule
Variant & Params (M) & F1 & EM & $\Delta$F1 (pp) & TTFT p50 / p90 (ms) \\
\midrule
Full Prefill & -- & 0.3155 & 0.232 & -- & 68.24 / 90.81 \\
Stale KV & -- & 0.1462 & 0.072 & -- & 22.41 / 24.33 \\
\midrule
A0 CacheRepair & 29.8 & 0.2462 & 0.172 & $0$ & 28.86 / 33.23 \\
A1 Token-causal & 29.8 & 0.2421 & 0.172 & $-0.41$ & 28.84 / 32.93 \\
A2 Bidirectional & 29.8 & 0.2427 & 0.164 & $-0.35$ & 29.42 / 33.48 \\
A3 Block diagonal & 29.8 & 0.0133 & 0.000 & $-23.29$ & 29.57 / 33.46 \\
A4 Entrance-only & 28.3 & 0.2400 & 0.162 & $-0.62$ & 28.68 / 32.98 \\
A5 Direct joint KV & 29.8 & 0.0049 & 0.000 & $-24.13$ & 29.12 / 32.95 \\
\bottomrule
\end{tabular}
\end{table}

\subsection{KV error before and after repair}
\label{sec:repair-mechanism}
We compare the 30M Qwen2.5-3B repairer's KV output with
stale and joint KV on MuSiQue. We pool squared error and joint-KV magnitude as in
Section~\ref{sec:observations}. In later chunks, the boundary region contains
the first eight tokens; the interior contains the remaining tokens.
Repair reduces K relative RMSE from 0.310 to 0.143 at boundaries
(53.9\%) and from 0.182 to 0.089 in interiors (50.9\%). V relative RMSE
decreases by 24.4\% and 16.2\%, respectively. Paired 95\% bootstrap intervals
for all four reductions lie above zero. Repair reduces error in both
boundary and interior regions.
Appendix~\ref{app:repair-mechanism} provides layer--position maps,
distance profiles, and regional uncertainty estimates.
These improvements also bring the target LLM's behavior closer to Full
Prefill. On 256 MuSiQue requests, repair reduces teacher-forced output KL
from 1.731 to 0.555 and query attention-output relative RMSE from 0.598
to 0.328 compared with Stale KV. Both metrics measure deviation from Full
Prefill (Appendix~\ref{app:training-objectives}).

\noindent\textbf{Sensitivity to chunk count.}
With document content fixed, increasing the chunk count
from 8 to 16 reduces Stale KV F1 from 0.231 to 0.165. CacheRepair 30M
achieves 0.289 and 0.277, respectively, while its p50 TTFT stays near
33\,ms. CacheRepair thus maintains higher answer quality than Stale KV
as chunk count increases, with nearly constant TTFT.
Appendix~\ref{app:scaling-amortization} reports all methods at
4, 8, 12, and 16 chunks, including paired quality intervals and resource
measurements.

\noindent\textbf{Computational advantage of all-token repair.}
At 16K document tokens, the largest repairers achieve
3.43--6.14$\times$ p50 TTFT speedups over Full Prefill while using
2.0--4.0\% of its dense matrix MACs. Across the measured 4K, 8K, and 16K
lengths, all three repairer capacities have lower p50 TTFT than every
CacheBlend and InfoFlow preset. Comparisons with fixed per-chunk EPIC
budgets depend on length and repairer capacity.
Appendix~\ref{app:scaling-amortization} reports all 270 configurations,
measured crossovers, and memory costs.

\begin{table}[htbp]
\centering
\caption{CacheRepair versus the highest-scoring baseline at half of Full Prefill p50 TTFT. Each method first selects its largest tested setting satisfying the budget, using latency alone. Brackets are simultaneous 95\% paired-bootstrap bands across all eligible baseline methods within that panel; complete pairwise comparisons accompany the artifact. A dash indicates no eligible repairer.}
\label{tab:budget-comparisons}
\footnotesize
\setlength{\tabcolsep}{4pt}
\begin{tabular}{llllr}
\toprule
Target & Dataset & Repair & Baseline & $\Delta$F1 (pp) and interval \\
\midrule
Qwen2.5-3B & MuSiQue & 30M & CacheBlend 0.1 & +5.80 [+1.45, +10.15] \\
Qwen2.5-3B & HotpotQA & -- & -- & -- \\
Qwen2.5-3B & MultiHop-RAG & 51M & KV Packet & -1.53 [-5.65, +2.60] \\
Qwen2.5-3B & TriviaQA & 51M & EPIC 32 & -0.68 [-4.46, +3.11] \\
Llama-3.1-8B & MuSiQue & 93M & CacheBlend 0.1 & +6.69 [+2.07, +11.31] \\
Llama-3.1-8B & HotpotQA & 93M & KV Packet & +10.21 [+5.49, +14.94] \\
Llama-3.1-8B & MultiHop-RAG & 93M & CacheBlend 0.1 & +1.07 [-3.18, +5.31] \\
Llama-3.1-8B & TriviaQA & 93M & CacheBlend 0.1 & +0.33 [-3.16, +3.82] \\
Qwen2.5-14B & MuSiQue & 159M & InfoFlow 0.1 & +3.93 [-0.65, +8.50] \\
Qwen2.5-14B & HotpotQA & 159M & EPIC 32 & +12.72 [+7.78, +17.67] \\
Qwen2.5-14B & MultiHop-RAG & 159M & InfoFlow 0.1 & +1.34 [-2.18, +4.85] \\
Qwen2.5-14B & TriviaQA & 159M & InfoFlow 0.1 & +1.36 [-1.57, +4.29] \\
\bottomrule
\end{tabular}
\end{table}

\section{Discussion}
\label{sec:discussion}

\noindent\textbf{Per-target-LLM training and deployment.}
Each frozen target LLM uses its own repairer and
normalization statistics (Table~\ref{tab:repair-scales}). This design targets
services with stable target LLMs and frequent document reuse. Training
and statistics costs equal the saved first-token time of 2.02--5.80 million
serial requests (Appendix~\ref{app:scaling-amortization}); actual service
costs depend on load and hardware. Changes to model weights, quantization,
or RoPE configuration require checking the resulting KV distribution and
repairer.

\noindent\textbf{Document length and serving conditions.}
The main Pareto curves measure single-request TTFT
on one A800 with warm pinned-host caches. All-token repair has
$\Theta(N^2)$ attention cost. Appendix~\ref{app:scaling-amortization} reports
longer-context quality, system-cost crossovers, and concurrent serving.
The concurrent-serving measurements use per-request repair.
Future work will batch repair across requests and pipeline H2D transfer,
repair, and KV writes to improve concurrent throughput.

\noindent\textbf{Training and cross-dataset generalization.}
CacheRepair learns KV residuals through offline training
on generic retrieval documents. Each target-specific checkpoint and its
statistics are fixed before evaluation and reused across four downstream
datasets, selected as described in Appendix~\ref{app:cohort-construction}.
Under the same two-epoch training budget, MSE achieves the highest mean
answer F1 among the three evaluated objectives
(Appendix~\ref{app:training-objectives}), supporting its use as the default
training objective.

\section{Conclusion}

CacheRepair learns all-token KV residual repair for a
frozen target LLM and reuses each repairer across four downstream datasets.
The largest repairers achieve 1.69--4.61$\times$ p50 TTFT speedups over
Full Prefill and improve Stale KV F1 by 2.1--26.1 percentage points.
CacheRepair contributes nondominated quality--TTFT points in eleven of
twelve model--dataset settings among the evaluated configurations,
extending the frontier alongside selective recomputation and learned cache fusion.

\clearpage
\section*{AI Use Disclosure}
We used OpenAI's GPT-5.6 Sol and GPT-6 Astra to support drafting, editing,
and language polishing, as well as research execution.
Research assistance included refining experimental plans, implementing
and debugging code, analyzing and
interpreting results, and preparing figures and tables. The authors
directed these activities and take responsibility for the correctness
of the experiments, the interpretation of the results, and the final manuscript, including all AI-assisted content.

\bibliographystyle{iclr2026_conference}
\bibliography{cacherepair}

\appendix
\clearpage
\section*{Appendix Contents}
{
\renewcommand{\arraystretch}{1.6}
\begin{tabularx}{\linewidth}{@{}Xr@{}}
\textbf{Section} & \textbf{Page} \\
\midrule
\hyperref[app:evaluation-details]{\ref*{app:evaluation-details}\quad Evaluation Details} & \pageref{app:evaluation-details} \\
\hspace*{1.5em}\hyperref[app:cohort-construction]{\ref*{app:cohort-construction}\quad \nameref*{app:cohort-construction}} & \pageref{app:cohort-construction} \\
\hspace*{1.5em}\hyperref[app:additional-requests]{\ref*{app:additional-requests}\quad \nameref*{app:additional-requests}} & \pageref{app:additional-requests} \\
\hyperref[app:cross-model-budget]{\ref*{app:cross-model-budget}\quad Common-Budget Results Across Target LLMs} & \pageref{app:cross-model-budget} \\
\hyperref[app:repair-mechanism]{\ref*{app:repair-mechanism}\quad Detailed KV Repair Measurements} & \pageref{app:repair-mechanism} \\
\hyperref[app:residual-capacity]{\ref*{app:residual-capacity}\quad Residual Structure and Repair Capacity} & \pageref{app:residual-capacity} \\
\hyperref[app:training-objectives]{\ref*{app:training-objectives}\quad Training Objectives and Functional Effects} & \pageref{app:training-objectives} \\
\hyperref[app:attention-topology]{\ref*{app:attention-topology}\quad Attention Topology and Within-Chunk Context} & \pageref{app:attention-topology} \\
\hyperref[app:observation-generality]{\ref*{app:observation-generality}\quad Cross-Model Validation of KV-Error Structure} & \pageref{app:observation-generality} \\
\hyperref[app:learned-baseline-compatibility]{\ref*{app:learned-baseline-compatibility}\quad Learned-Fusion Baseline Compatibility} & \pageref{app:learned-baseline-compatibility} \\
\hyperref[app:scaling-amortization]{\ref*{app:scaling-amortization}\quad Scaling and Amortization Details} & \pageref{app:scaling-amortization} \\
\hyperref[app:generic-corpus]{\ref*{app:generic-corpus}\quad Generic Repair Pretraining Corpus Composition} & \pageref{app:generic-corpus} \\
\end{tabularx}
}
\clearpage

\section{Evaluation Details}
\label{app:evaluation-details}

\noindent\textbf{Comparison scope.}
We evaluate cache fusion with a fixed target LLM, preserving its deployed
weights and behavior on requests that do not use document-cache fusion.
Methods that fine-tune the target LLM address a complementary setting
with a different joint-prefill reference. Our baselines share the frozen
target and downstream inputs, allowing the comparison to isolate how
each method constructs the reusable document KV.

\noindent\textbf{Repairer configurations.}
Table~\ref{tab:repair-scales} reports the three capacity levels for each
target. Width $d_b$, block count $B$, and segment dimension
$d_{\mathrm{seg}}$ determine capacity together with the target-specific
encoder, token projection, and KV head. Each Repair Block has
64-dimensional attention heads and an MLP expansion ratio of three.
Pareto points are labeled by rounded trainable parameter counts.

\begin{table}[ht]
    \centering
    \caption{Trained CacheRepair configurations for each target LLM. Parameter counts are rounded to the nearest million.}
    \label{tab:repair-scales}
    \small
    \setlength{\tabcolsep}{4.5pt}
    {
    \begin{tabular}{llrrrr}
        \toprule
        Target LLM & Size & $d_b$ & $B$ & $d_{\mathrm{seg}}$ & Parameters \\
        \midrule
        Qwen2.5-3B & Small  & 256 & 5 & 8  & 9M \\
                   & Medium & 512 & 6 & 16 & 30M \\
                   & Large  & 704 & 6 & 24 & 51M \\
        \midrule
        Llama-3.1-8B & Small  & 256 & 5 & 8  & 23M \\
                    & Medium & 512 & 6 & 16 & 59M \\
                    & Large  & 704 & 6 & 24 & 93M \\
        \midrule
        Qwen2.5-14B & Small  & 320 & 5 & 8  & 42M \\
                    & Medium & 640 & 6 & 16 & 104M \\
                    & Large  & 832 & 7 & 24 & 159M \\
        \bottomrule
    \end{tabular}}
\end{table}

\noindent\textbf{Document preparation.}
MultiHop-RAG uses query-only BM25 retrieval, with the contexts partitioned
into 20 chunks; TriviaQA also uses 20 chunks. MuSiQue and HotpotQA retain
their fixed context segmentation. We preserve context order and content
tokenization. Each target's native chat template wraps the context and full question,
prepending opening tokens to the first chunk. After the documents,
the user message contains two newline characters followed by the exact suffix below;
\texttt{<question>} is replaced with the full question. The native chat template
then supplies the assistant-generation prefix.
\begin{quote}
\small
Answer the question based on the passages. Return only the minimal answer phrase,
with no explanation or extra description. Do not restate the question.\\
Question: \texttt{<question>}\\
Answer:
\end{quote}
The 4,096-token document budget yields prefixes
of at most 4,120 tokens and query suffixes of at most 158 tokens.

\noindent\textbf{Online execution and timing.}
We use vLLM~0.8.5 V1, BF16, tensor parallelism one, and greedy generation
capped at 32 tokens. Online TTFT accounts for method-specific H2D transfer,
global RoPE, repair or scoring/selection/recomputation, query processing,
and first-token generation. Independent chunk compilation and pinned-host
preparation precede timing. For InfoFlow, separately timed scoring,
selection, and main-preparation costs are added to the main generation's
scheduler-to-first-token measurement, including work before scheduler
admission. Stale KV and KV Packet share the cached-prefix transfer,
global-RoPE, and paged-cache injection implementation. Stale KV and CacheRepair
load independently compiled and repaired prefixes, respectively, through the
last complete 16-token cache block. Any remaining document tokens are
processed with the query and included in TTFT. KV Packet loads
its complete physical prefix, including learned header and trailer entries.

\noindent\textbf{Optimization.}
{
We use AdamW with $(\beta_1,\beta_2)=(0.9,0.95)$, weight decay 0.01, unit
gradient clipping, an effective document batch of four, and BF16 target-LLM
execution. The main repairers follow a 100,000-update learning-rate schedule:
2,000 warmup updates to $3\times10^{-4}$, followed by cosine decay toward
$3\times10^{-5}$. We evaluate update 75,000, after six complete passes over
the training corpus. The controlled two-epoch comparisons use a
25,000-update schedule with 500 warmup updates and the same peak and final
learning rates.}

\noindent\textbf{Recomputation grids.}
CacheBlend uses its released V-deviation selector at ratios 0.10, 0.20, 0.40, and 0.60;
EPIC uses boundary widths 8, 16, 32, and 64; InfoFlow uses ratios 0.05, 0.10,
0.15, and 0.25 with attention-mass scores from layers
22--25 (zero-indexed), following the layer range used across models in
the original paper~\citep{teng2026infoflow}.
We retain this common preset across targets; the comparison measures
the released layer choice rather than a separately optimized selector for each LLM.
A scoring pass supplies a global token ranking;
selected tokens are then recomputed in their original positions. The scoring
pass ends after layer 25; the main generation executes every target-LLM
layer. Both passes reuse the request's materialized GPU cache.

\noindent\textbf{Serving and statistics.}
We score the first generated answer line using
conversion to lowercase, punctuation and article removal, and whitespace normalization;
standalone yes/no answers followed by punctuation are parsed as boolean
answers. F1 and EM take the maximum over the supplied gold aliases.
Mean-F1 confidence intervals use 10,000 request-bootstrap draws with seed
20260730; differences between methods use paired draws of the same requests.

\noindent\textbf{Full recomputation.}
On all 500 Qwen2.5-3B/MultiHop-RAG requests and all 500
Llama-3.1-8B/TriviaQA requests, setting CacheBlend or InfoFlow to recompute
every document token reproduces Full Prefill's generated tokens exactly.
Full Prefill also produces identical tokens using the baseline engine's
cache allocation and batching limits, and through its evaluation entry
point. All four comparisons preserve inputs and decoding settings;
all 4,000 generations match Full Prefill.

\subsection{Dataset selection}
\label{app:cohort-construction}
We select 500 requests for each downstream dataset from the complete public
source pool: MuSiQue development, HotpotQA validation (distractor),
MultiHop-RAG's public questions, and TriviaQA validation (reading
comprehension). Selection uses a fixed comparison against the original
passage texts referenced by the actual repair-training manifests.
The three manifests share the same 50,000 source-query rows and ordered
passage texts, yielding 488,823 distinct passage texts. We normalize both
sides with Unicode NFKC, case folding, and alphanumeric tokens separated at
punctuation and whitespace, retaining numbers and stopwords.

\noindent\textbf{Matching rules.}
A request is excluded when any of the following conditions holds.
\emph{A (evidence sentence):} a complete evidence sentence of at least eight
normalized tokens occurs contiguously in a training passage. Evidence uses
HotpotQA's supporting sentences, sentences of MuSiQue's supporting
paragraphs, MultiHop-RAG's evidence facts, and TriviaQA's supplied document
evidence. Sentence splitting uses English PySBD~0.3.4.
\emph{B (input context):} a contiguous 50-token span in any supplied context
document matches a training passage. Both sides use stride one; context
coverage includes distractors and target-specific retained text, with
source-document boundaries preserved. Hash lookups are followed by exact
token comparisons.
\emph{C (question provenance):} a normalized full or resolved component
question equals a training-source question, or a preserved source-ID mapping
or confirmed question/fact correspondence identifies the training seed.
MuSiQue component references are resolved using their preceding component
answers; the mapping records include the official single-hop identifiers.

\noindent\textbf{Deterministic selection and shared inputs.}
We rank source IDs by SHA256 of the fixed string
\texttt{20260908|dataset|source\_id}, remove repeated normalized questions,
and take the first 500 eligible requests plus 50 reserves. Screening stops
after these 550 requests; Table~\ref{tab:cohort-construction} reports the
candidates examined. Model predictions are not used for selection.
All selected requests are checked again under the same rules. The four
downstream datasets contain 2,000 distinct normalized questions. All target LLMs and
methods use the same sample identities and order, with fixed retrieved
content, target-specific tokenization, and native chat templates. Context
selection uses question text and source context order; reference answers
are used for scoring. Each measured request supplies both quality and TTFT.

\begin{table}[ht]
\centering
\caption{Request selection for the four downstream datasets.
Rule counts refer to examined candidates and can overlap; the union column
counts requests excluded by at least one rule. Each dataset retains 500
evaluation requests and 50 reserves.}
\label{tab:cohort-construction}
\small
\setlength{\tabcolsep}{5pt}
{
\begin{tabular}{lrrrrrrr}
\toprule
Dataset & Public pool & Examined & A & B & C & Union & Repeated \\
\midrule
MuSiQue & 2,417 & 2,375 & 705 & 1,695 & 264 & 1,818 & 7 \\
HotpotQA & 7,405 & 822 & 67 & 249 & 0 & 272 & 0 \\
MultiHop-RAG & 2,556 & 550 & 0 & 0 & 0 & 0 & 0 \\
TriviaQA & 17,944 & 1,123 & 540 & 379 & 1 & 541 & 32 \\
\bottomrule
\end{tabular}}
\end{table}

The code release provides the fixed evaluation question IDs, input hashes,
source specifications, and scripts for reconstructing the evaluation inputs. The checks cover the stated text and provenance conditions;
semantic paraphrases and target-LLM pretraining corpora remain outside their scope.

\noindent\textbf{MuSiQue composition.}
All methods are compared on the same 500 questions, selected by fixed
content rules before generation. The selection spans two-, three-, and
four-hop reasoning.
Using the official source IDs to identify hop count, the public pool has
1,252/760/405 two-/three-/four-hop questions (51.8/31.4/16.8\%).
The selected 500 contain 283/179/38 (56.6/35.8/7.6\%); the 2,375 examined
candidates have proportions 51.8/31.5/16.7\%.
These counts specify the evaluation mixture used for every method and
make its composition comparable with the public source pool.

\subsection{Answer-F1 variation}
\label{app:additional-requests}
\label{app:answer-comparisons}
Full Prefill provides the fully contextualized KV reference.
Cache fusion changes the generated answer, including its content and wording;
its answer F1 is therefore not constrained to lie between Stale KV and Full
Prefill. When the reference scores are close---as on
Qwen2.5-3B/MultiHop-RAG, where the original-batch gap is 2.77 percentage points---changes
on a small number of requests can alter the ordering of mean scores.

We select 500 additional requests for each of Qwen2.5-3B/MultiHop-RAG
and Llama-3.1-8B/TriviaQA using Appendix~\ref{app:cohort-construction}'s
rules, freezing both lists before generation. All 18 configurations retain
their weights and evaluation settings. We compare the original, additional,
and combined batches using 10,000 paired bootstrap draws (seed 20260911);
combined TTFT pools all per-request measurements.

For Qwen2.5-3B/MultiHop-RAG, CacheRepair 30M minus KV Packet
is $-1.51$ F1 points [$-4.98$, $2.01$] on the original 500 requests and
$+3.82$ [$0.22$, $7.42$] on the additional 500. Across all 1,000 requests,
their F1 scores are 55.20\% and 54.04\%, with a paired difference of
$+1.16$ [$-1.32$, $3.62$]; their p50 TTFTs are 41.29 and 27.78\,ms, respectively.
Thus the batches give different quality orderings while KV Packet remains
faster. On these combined requests,
KV Packet and CacheBlend 0.6 differ from Full Prefill by $-0.84$ and
$+1.34$ points; their pointwise 95\% intervals both include zero.
On Llama-3.1-8B/TriviaQA, InfoFlow 0.15 scores higher than Full Prefill in both batches,
with a combined difference of $+2.18$ points [$0.88$, $3.51$].
Table~\ref{tab:answer-level-comparisons} gives the original-batch paired results. Answer wording contributes: Full Prefill produces ``Ozone (a form of oxygen)'' and InfoFlow ``Ozone.'', scoring 0.40 and 1.00 token F1, respectively. Other changed predictions identify different entities.

\begin{table}[htbp]
\centering
\caption{Answer-level comparisons on the complete 500-request datasets. F1 is shown as a percentage. Differences are method minus Full Prefill, with pointwise 95\% paired request-bootstrap intervals from 10,000 draws. W/L counts requests with higher/lower F1 than Full Prefill; the remaining requests have equal F1.}
\label{tab:answer-level-comparisons}
\footnotesize
\setlength{\tabcolsep}{4pt}
{\begin{tabular}{llrrrr}
\toprule
Target / dataset & Method & Full F1 & Method F1 & $\Delta$F1 (pp) & W/L \\
\midrule
\shortstack[l]{Qwen2.5-3B\\MultiHop-RAG} & CacheBlend 0.6 & 54.28 & 56.26 & +1.98 [-0.023, +4.054] & 19/10 \\
\shortstack[l]{Qwen2.5-3B\\MultiHop-RAG} & KV Packet 8+8 & 54.28 & 55.09 & +0.81 [-2.867, +4.447] & 46/43 \\
\shortstack[l]{Llama-3.1-8B\\TriviaQA} & CacheRepair 93M & 80.49 & 82.30 & +1.81 [-0.001, +3.655] & 34/19 \\
\shortstack[l]{Llama-3.1-8B\\TriviaQA} & InfoFlow 0.15 & 80.49 & 82.94 & +2.45 [+0.556, +4.324] & 41/17 \\
\bottomrule
\end{tabular}}
\end{table}

\section{Common-Budget Results Across Target LLMs}
\label{app:cross-model-budget}

Tables~\ref{tab:qwen3b-common-budget}, \ref{tab:llama8-common-budget}, and
\ref{tab:qwen14-common-budget} use $B=0.5T_{\mathrm{full}}$, selecting each method's largest tested
capacity or recomputation setting within the latency budget. The rule is
fixed across datasets and target LLMs and does not use answer quality.

We also compare the largest repairer for each target with
the highest tested recomputation presets: CacheBlend 0.6, EPIC 64, and
InfoFlow 0.25. Across all twelve model--dataset settings, CacheRepair has
lower p50 TTFT in all 36 comparisons, giving 1.30--6.76$\times$ speedups.
F1 differences span $-3.86$ to $+4.43$ percentage points.
Table~\ref{tab:high-budget-comparisons} reports all 36 paired differences
and pointwise 95\% intervals alongside the speedups. Together with the
half-latency results, these comparisons show how offline repair provides
lower-latency operating points alongside high-budget recomputation,
with the quality difference quantified for every setting.

\begin{table}[htbp]
\centering
\caption{Largest repairer versus each highest tested recomputation preset on the same 500 requests per dataset. Each cell gives repair minus baseline F1 in percentage points, its pointwise 95\% paired bootstrap interval, and baseline/repair p50 TTFT (speedup). Intervals are not adjusted jointly across the 36 comparisons.}
\label{tab:high-budget-comparisons}
\scriptsize
\setlength{\tabcolsep}{5pt}
\renewcommand{\arraystretch}{1.12}
\begin{tabular}{lccc}
\toprule
Dataset & CacheBlend 0.6 & EPIC64 & InfoFlow 0.25 \\
\midrule
\multicolumn{4}{l}{\textit{Qwen2.5-3B / 51M}} \\
MuSiQue & \makecell{+0.98 [-2.38, +4.35] \\ 2.13$\times$} & \makecell{-1.11 [-4.41, +2.17] \\ 1.84$\times$} & \makecell{+3.19 [-0.18, +6.56] \\ 1.80$\times$} \\
HotpotQA & \makecell{+3.74 [+0.63, +6.90] \\ 1.72$\times$} & \makecell{+3.83 [+0.65, +7.09] \\ 1.30$\times$} & \makecell{+4.43 [+1.19, +7.79] \\ 1.79$\times$} \\
MultiHop-RAG & \makecell{-2.69 [-5.00, -0.47] \\ 3.67$\times$} & \makecell{-1.05 [-3.66, +1.55] \\ 2.14$\times$} & \makecell{-0.50 [-3.00, +1.93] \\ 2.38$\times$} \\
TriviaQA & \makecell{-0.69 [-3.22, +1.92] \\ 3.53$\times$} & \makecell{+0.23 [-2.53, +3.00] \\ 2.05$\times$} & \makecell{+0.53 [-2.12, +3.20] \\ 2.33$\times$} \\
\midrule
\multicolumn{4}{l}{\textit{Llama-3.1-8B / 93M}} \\
MuSiQue & \makecell{-0.86 [-3.89, +2.13] \\ 3.41$\times$} & \makecell{-1.74 [-4.84, +1.44] \\ 3.05$\times$} & \makecell{-2.73 [-5.84, +0.35] \\ 2.29$\times$} \\
HotpotQA & \makecell{-3.82 [-6.65, -0.87] \\ 2.74$\times$} & \makecell{-0.23 [-3.34, +2.96] \\ 2.14$\times$} & \makecell{-1.23 [-4.14, +1.78] \\ 2.24$\times$} \\
MultiHop-RAG & \makecell{-0.73 [-3.44, +1.94] \\ 5.66$\times$} & \makecell{+2.14 [-0.43, +4.81] \\ 3.28$\times$} & \makecell{+1.08 [-1.58, +3.86] \\ 3.18$\times$} \\
TriviaQA & \makecell{+0.46 [-1.60, +2.56] \\ 5.29$\times$} & \makecell{+0.93 [-1.32, +3.17] \\ 3.18$\times$} & \makecell{-0.22 [-2.32, +1.90] \\ 3.02$\times$} \\
\midrule
\multicolumn{4}{l}{\textit{Qwen2.5-14B / 159M}} \\
MuSiQue & \makecell{-3.08 [-6.25, +0.23] \\ 4.08$\times$} & \makecell{-3.86 [-7.25, -0.50] \\ 3.45$\times$} & \makecell{-1.27 [-4.53, +2.07] \\ 2.43$\times$} \\
HotpotQA & \makecell{-3.66 [-6.71, -0.66] \\ 3.15$\times$} & \makecell{+0.86 [-2.39, +4.12] \\ 2.39$\times$} & \makecell{+0.57 [-2.38, +3.59] \\ 2.32$\times$} \\
MultiHop-RAG & \makecell{+0.16 [-1.95, +2.26] \\ 6.76$\times$} & \makecell{+1.52 [-1.07, +4.16] \\ 3.62$\times$} & \makecell{-0.67 [-3.02, +1.71] \\ 3.47$\times$} \\
TriviaQA & \makecell{+0.60 [-1.56, +2.69] \\ 6.43$\times$} & \makecell{+0.05 [-1.87, +1.96] \\ 3.53$\times$} & \makecell{+0.69 [-1.25, +2.57] \\ 3.30$\times$} \\
\bottomrule
\end{tabular}
\end{table}

\begin{table}[t]
\centering
\caption{Qwen2.5-3B quality at a TTFT budget of $B=0.5T_{\mathrm{full}}$. Each method uses its largest tested setting that meets the budget. Cells report F1/EM and p50 TTFT normalized by Full Prefill. Full Prefill and Stale KV provide reference values; a dash indicates no eligible setting.}
\label{tab:qwen3b-common-budget}
\scriptsize
\setlength{\tabcolsep}{3.2pt}
{\begin{tabular}{lcccccccc}
\toprule
 & \multicolumn{2}{c}{MuSiQue} & \multicolumn{2}{c}{HotpotQA} & \multicolumn{2}{c}{MultiHop-RAG} & \multicolumn{2}{c}{TriviaQA} \\
\cmidrule(lr){2-3}\cmidrule(lr){4-5}\cmidrule(lr){6-7}\cmidrule(lr){8-9}
Method & F1 / EM & TTFT / Full & F1 / EM & TTFT / Full & F1 / EM & TTFT / Full & F1 / EM & TTFT / Full \\
\midrule
Full Prefill & 0.316 / 0.232 & 1.000 & 0.566 / 0.430 & 1.000 & 0.543 / 0.540 & 1.000 & 0.799 / 0.708 & 1.000 \\
Stale KV & 0.146 / 0.072 & 0.328 & 0.366 / 0.256 & 0.438 & 0.515 / 0.506 & 0.217 & 0.707 / 0.622 & 0.224 \\
CacheBlend & 0.198 / 0.120 & 0.463 & -- & -- & 0.534 / 0.528 & 0.416 & 0.773 / 0.694 & 0.415 \\
EPIC & 0.182 / 0.112 & 0.461 & 0.400 / 0.284 & 0.471 & 0.545 / 0.540 & 0.461 & 0.776 / 0.700 & 0.473 \\
InfoFlow & -- & -- & -- & -- & 0.539 / 0.532 & 0.496 & 0.774 / 0.690 & 0.489 \\
KV Packet & 0.191 / 0.126 & 0.345 & 0.394 / 0.278 & 0.461 & 0.551 / 0.544 & 0.218 & 0.723 / 0.636 & 0.237 \\
CacheRepair & 0.256 / 0.170 & 0.488 & -- & -- & 0.536 / 0.532 & 0.336 & 0.769 / 0.688 & 0.350 \\
\bottomrule
\end{tabular}}
\end{table}

\begin{table}[htbp]
\centering
\caption{Llama-3.1-8B quality at a TTFT budget of $B=0.5T_{\mathrm{full}}$. Each method uses its largest tested setting that meets the budget. Cells report F1/EM and p50 TTFT normalized by Full Prefill. Full Prefill and Stale KV provide reference values; a dash indicates no eligible setting.}
\label{tab:llama8-common-budget}
\scriptsize
\setlength{\tabcolsep}{3.2pt}
{\begin{tabular}{lcccccccc}
\toprule
 & \multicolumn{2}{c}{MuSiQue} & \multicolumn{2}{c}{HotpotQA} & \multicolumn{2}{c}{MultiHop-RAG} & \multicolumn{2}{c}{TriviaQA} \\
\cmidrule(lr){2-3}\cmidrule(lr){4-5}\cmidrule(lr){6-7}\cmidrule(lr){8-9}
Method & F1 / EM & TTFT / Full & F1 / EM & TTFT / Full & F1 / EM & TTFT / Full & F1 / EM & TTFT / Full \\
\midrule
Full Prefill & 0.395 / 0.284 & 1.000 & 0.680 / 0.510 & 1.000 & 0.683 / 0.660 & 1.000 & 0.805 / 0.692 & 1.000 \\
Stale KV & 0.205 / 0.126 & 0.263 & 0.439 / 0.312 & 0.331 & 0.453 / 0.442 & 0.208 & 0.735 / 0.664 & 0.212 \\
CacheBlend & 0.285 / 0.204 & 0.494 & -- & -- & 0.643 / 0.630 & 0.494 & 0.820 / 0.734 & 0.473 \\
EPIC & 0.266 / 0.186 & 0.491 & 0.492 / 0.362 & 0.478 & 0.603 / 0.592 & 0.428 & 0.813 / 0.724 & 0.388 \\
InfoFlow & -- & -- & -- & -- & 0.627 / 0.614 & 0.498 & 0.815 / 0.730 & 0.458 \\
KV Packet & 0.246 / 0.154 & 0.278 & 0.509 / 0.380 & 0.342 & 0.576 / 0.562 & 0.215 & 0.773 / 0.666 & 0.215 \\
CacheRepair & 0.352 / 0.262 & 0.380 & 0.612 / 0.460 & 0.421 & 0.653 / 0.634 & 0.287 & 0.823 / 0.724 & 0.291 \\
\bottomrule
\end{tabular}}
\end{table}

\begin{table}[htbp]
\centering
\caption{Qwen2.5-14B quality at a TTFT budget of $B=0.5T_{\mathrm{full}}$. Each method uses its largest tested setting that meets the budget. Cells report F1/EM and p50 TTFT normalized by Full Prefill. Full Prefill and Stale KV provide reference values; a dash indicates no eligible setting.}
\label{tab:qwen14-common-budget}
\scriptsize
\setlength{\tabcolsep}{3.2pt}
{\begin{tabular}{lcccccccc}
\toprule
 & \multicolumn{2}{c}{MuSiQue} & \multicolumn{2}{c}{HotpotQA} & \multicolumn{2}{c}{MultiHop-RAG} & \multicolumn{2}{c}{TriviaQA} \\
\cmidrule(lr){2-3}\cmidrule(lr){4-5}\cmidrule(lr){6-7}\cmidrule(lr){8-9}
Method & F1 / EM & TTFT / Full & F1 / EM & TTFT / Full & F1 / EM & TTFT / Full & F1 / EM & TTFT / Full \\
\midrule
Full Prefill & 0.450 / 0.348 & 1.000 & 0.733 / 0.580 & 1.000 & 0.731 / 0.722 & 1.000 & 0.837 / 0.740 & 1.000 \\
Stale KV & 0.176 / 0.120 & 0.202 & 0.388 / 0.288 & 0.247 & 0.675 / 0.672 & 0.165 & 0.774 / 0.686 & 0.172 \\
CacheBlend & 0.277 / 0.218 & 0.407 & 0.461 / 0.348 & 0.395 & 0.696 / 0.686 & 0.408 & 0.794 / 0.698 & 0.410 \\
EPIC & 0.284 / 0.214 & 0.406 & 0.522 / 0.400 & 0.488 & 0.697 / 0.688 & 0.492 & 0.800 / 0.704 & 0.328 \\
InfoFlow & 0.338 / 0.254 & 0.489 & 0.498 / 0.376 & 0.480 & 0.714 / 0.708 & 0.461 & 0.808 / 0.716 & 0.466 \\
KV Packet & 0.275 / 0.198 & 0.214 & 0.499 / 0.384 & 0.252 & 0.694 / 0.688 & 0.172 & 0.790 / 0.704 & 0.179 \\
CacheRepair & 0.378 / 0.282 & 0.289 & 0.649 / 0.510 & 0.313 & 0.727 / 0.720 & 0.217 & 0.822 / 0.732 & 0.229 \\
\bottomrule
\end{tabular}}
\end{table}

\noindent\textbf{Quality and budget comparisons.}
Table~\ref{tab:quality-tradeoffs} pairs the largest
repairer's speedup with its F1 difference from Full Prefill.
Table~\ref{tab:budget-comparisons} compares the fixed presets selected by
the half-latency budget. Within each panel, the intervals cover all
eligible baseline comparisons simultaneously using the maximum absolute
centered error over 10,000 paired bootstrap draws. CacheRepair's mean
exceeds every eligible baseline in nine panels; the simultaneous intervals
remain positive in four: Qwen2.5-3B/MuSiQue, Llama-3.1-8B/MuSiQue,
Llama-3.1-8B/HotpotQA, and Qwen2.5-14B/HotpotQA.

\begin{table}[htbp]
\centering
\caption{Answer quality and latency of the largest repairer for each target. F1 is in percent; differences use paired 95\% request-bootstrap intervals on each complete 500-request dataset. Speedup is Full Prefill p50 TTFT divided by repair p50 TTFT.}
\label{tab:quality-tradeoffs}
\scriptsize
\setlength{\tabcolsep}{4pt}
{\begin{tabular}{lllrrrr}
\toprule
Target & Dataset & Capacity & Full F1 & Repair F1 & $\Delta$F1 (pp) & Speedup \\
\midrule
Qwen2.5-3B & MuSiQue & 51M & 31.55 & 26.26 & -5.29 [-8.51, -2.00] & 1.96 \\
Qwen2.5-3B & HotpotQA & 51M & 56.61 & 55.35 & -1.26 [-3.91, +1.39] & 1.69 \\
Qwen2.5-3B & MultiHop-RAG & 51M & 54.28 & 53.57 & -0.71 [-2.95, +1.44] & 2.98 \\
Qwen2.5-3B & TriviaQA & 51M & 79.85 & 76.94 & -2.91 [-5.15, -0.73] & 2.86 \\
Llama-3.1-8B & MuSiQue & 93M & 39.48 & 35.18 & -4.30 [-7.18, -1.39] & 2.63 \\
Llama-3.1-8B & HotpotQA & 93M & 67.98 & 61.16 & -6.82 [-9.68, -4.08] & 2.38 \\
Llama-3.1-8B & MultiHop-RAG & 93M & 68.33 & 65.33 & -3.00 [-5.83, -0.23] & 3.48 \\
Llama-3.1-8B & TriviaQA & 93M & 80.49 & 82.30 & +1.81 [-0.00, +3.65] & 3.44 \\
Qwen2.5-14B & MuSiQue & 159M & 44.99 & 37.78 & -7.21 [-10.23, -4.24] & 3.46 \\
Qwen2.5-14B & HotpotQA & 159M & 73.32 & 64.93 & -8.38 [-11.23, -5.62] & 3.19 \\
Qwen2.5-14B & MultiHop-RAG & 159M & 73.06 & 72.71 & -0.35 [-2.57, +1.82] & 4.61 \\
Qwen2.5-14B & TriviaQA & 159M & 83.74 & 82.21 & -1.53 [-3.42, +0.37] & 4.38 \\
\bottomrule
\end{tabular}}
\end{table}

\FloatBarrier
\section{Detailed KV Repair Measurements}
\label{app:repair-mechanism}
\begin{figure}[ht]
\centering
\includegraphics[width=\linewidth]{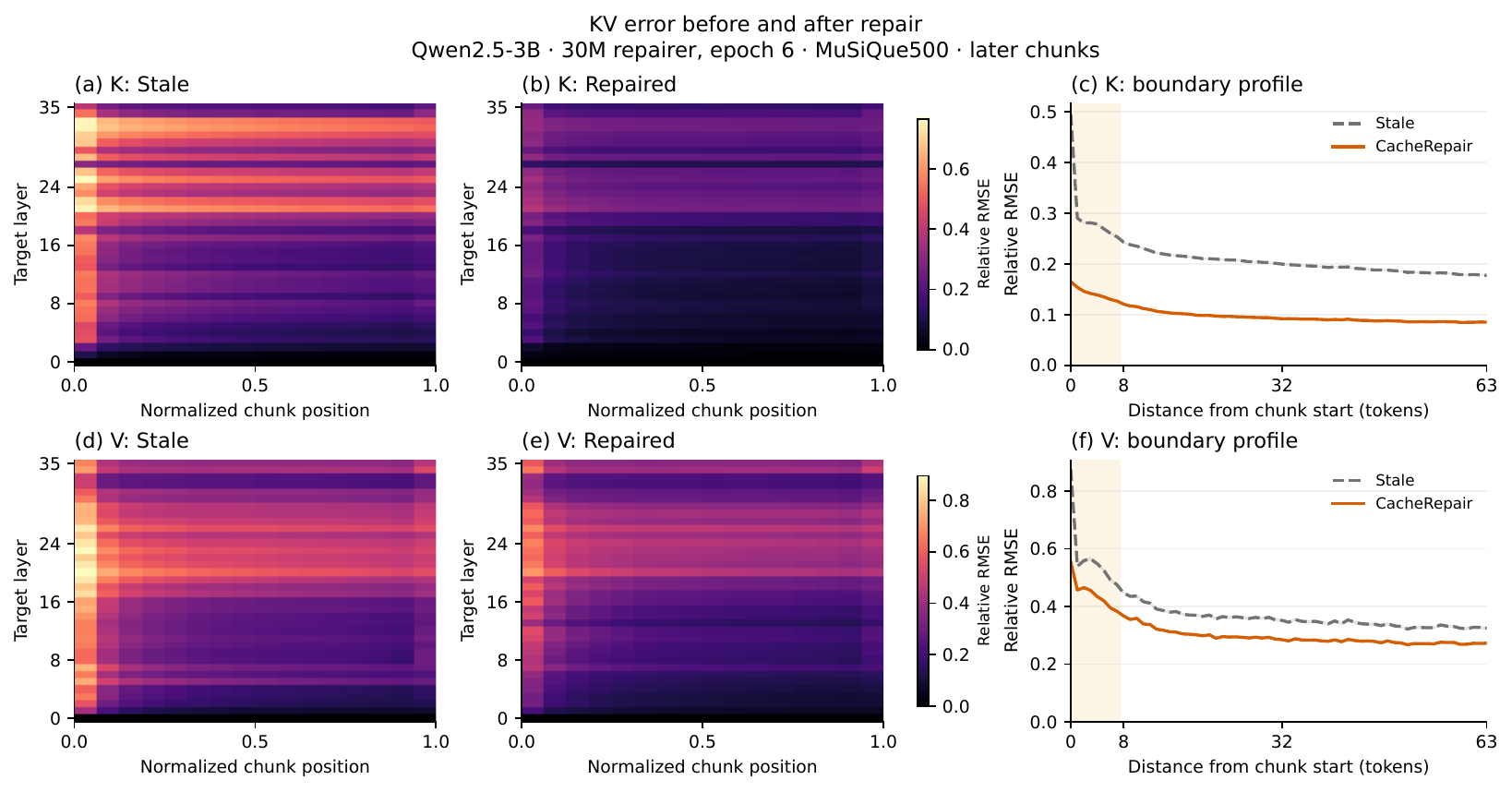}
\caption{KV error before and after epoch-6 repair on
Qwen2.5-3B (500 MuSiQue requests). K and V occupy separate rows; columns
show stale error, repaired error, and error by distance from chunk start.
Heatmap scales are shared within each row. Shading on the profiles shows
95\% request-bootstrap intervals.}
\label{fig:repair-mechanism}
\end{figure}

Figure~\ref{fig:repair-mechanism} shows the epoch-6
30M repairer's error on 500 MuSiQue requests with Qwen2.5-3B. Heatmaps
pool later-chunk tokens into 16 normalized position bins. Distance profiles
pool layers and heads at each token offset from a later chunk's start.
Uncertainty uses 10,000 paired request-bootstrap draws with seed 20260730.
The 95\% intervals for relative error reduction are 53.7--54.1\% (K boundary),
50.5--51.2\% (K interior), 24.2--24.6\% (V boundary), and
15.9--16.5\% (V interior). Across all document tokens, relative RMSE
decreases by 51.4\% for K and 18.0\% for V.
For the first chunk, K relative RMSE is 0.0068/0.0097 for stale/repaired
KV, and V relative RMSE is 0.0268/0.0294.
These measurements characterize the epoch-6 repairer; controlled comparisons
of individual design choices use separately trained models.

\section{Residual Structure and Repair Capacity}
\label{app:residual-capacity}
\begin{figure}[htbp]
\centering
\includegraphics[width=\linewidth]{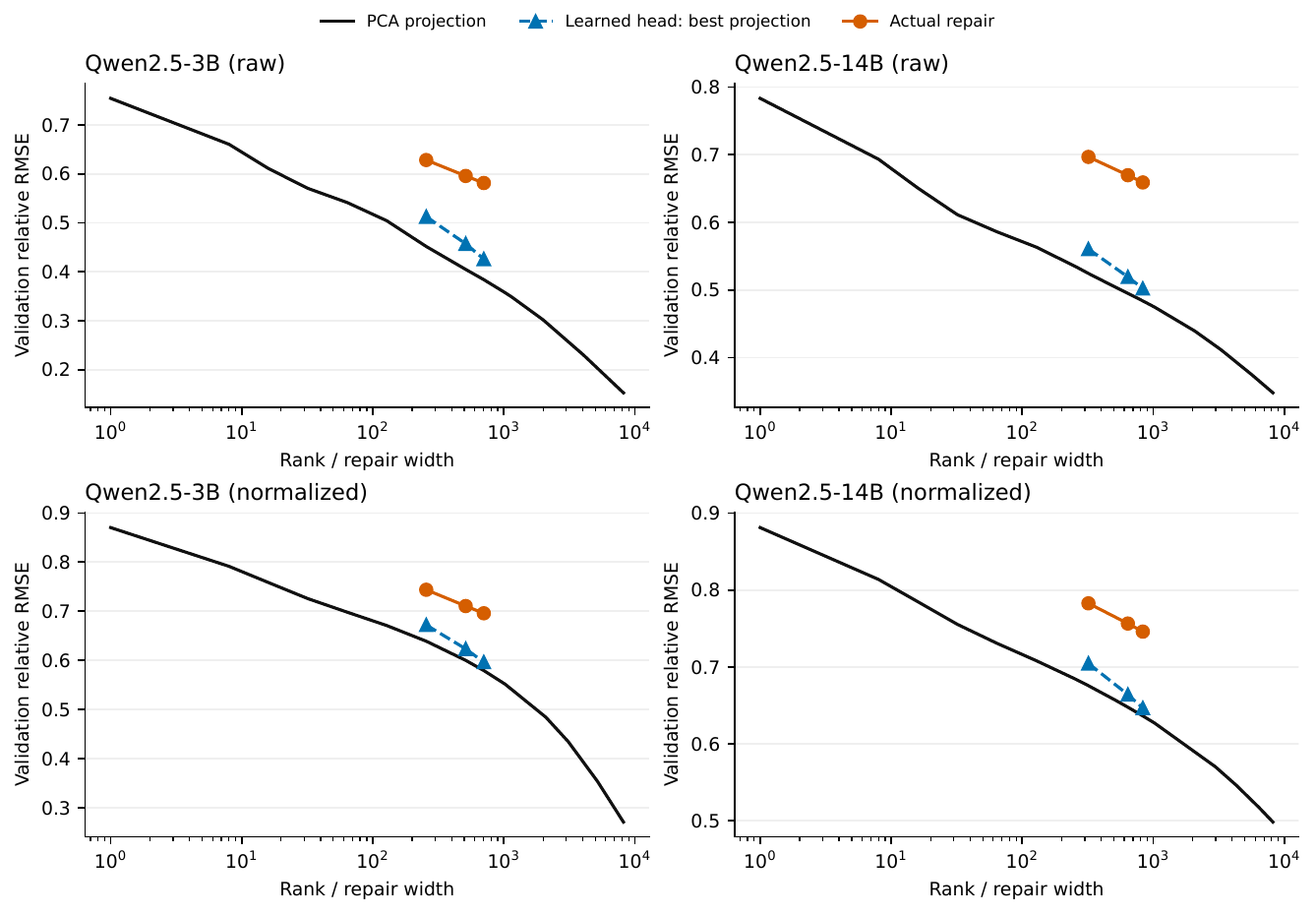}
\caption{Residual relative RMSE on the separate set of 128
validation requests. PCA uses a basis and mean from the fitting requests;
the learned-head projection uses each checkpoint's fixed affine output
space. Actual repair uses the corresponding checkpoint's predictions.}
\label{fig:residual-validation}
\end{figure}

{
\noindent\textbf{Residual representation.}
We use the first 256 MuSiQue requests from Section~\ref{sec:observations},
taking 64 evenly spaced document positions per request.
We use the first 128 requests to fit the mean and PCA bases,
and the remaining 128 to evaluate projections and the fixed epoch-six repairers.
Each token's canonical-K/native-V residual is flattened over layers and
heads, giving 18,432 coordinates for Qwen2.5-3B and 98,304 for Qwen2.5-14B.
We analyze raw residuals and residuals divided by the fixed training
normalization scales. Centering uses the fitting mean in both splits;
the fitting matrix has rank at most 8,191. Spectra and projections use FP64
arithmetic on the stored FP32 residuals.

The residual spectra quantify the available projection capacity. Raw residuals require ranks 1,990 and 3,298
to retain 95\% of fitting energy for Qwen2.5-3B and Qwen2.5-14B;
the normalized ranks are 3,105 and 4,282. At the three evaluated repair
widths, retained raw energy is 73.0/81.2/84.8\% for Qwen2.5-3B and
67.1/75.3/78.5\% for Qwen2.5-14B. The accompanying data report separate K/V and per-layer spectra.

\noindent\textbf{Projection capacity and prediction.}
For validation residuals $R_i$, we report
$\sqrt{\sum_i\lVert R_i-\widehat R_i\rVert_2^2/
\sum_i\lVert R_i\rVert_2^2}$ in the selected coordinates.
The PCA reference projects onto the fitted basis around the fitting mean.
The learned-head reference projects onto the fixed affine space
$b+\operatorname{col}(W_{\mathrm{out}})$. Both projections access the true
validation residual; actual repair predicts it from stale KV and token
features. Figure~\ref{fig:residual-validation} compares these three errors.
The learned output heads have rank $d_b$. Increasing capacity reduces
actual normalized residual MSE from 0.514 to 0.469 to 0.450 for
Qwen2.5-3B, and from 0.556 to 0.519 to 0.505 for Qwen2.5-14B.
These curves separate output-space capacity from residual prediction error.
The gap between the learned-head projection and actual repair shows that
prediction contributes error even within the available output space.
The projection has access to the true joint residual, whereas the repairer
must infer it from stale KV and token features; this gap does not by itself
identify which cross-chunk dependencies are hardest to recover.

\noindent\textbf{Parameter allocation.}
Table~\ref{tab:repair-parameter-groups} breaks down encoder, token-fusion,
reinjection, backbone, and output-head parameters. The output head scales
with the target LLM's KV dimension and accounts for 63.01M of the 104M
Qwen2.5-14B repairer's parameters.
For repair width $d_b$, the dense head uses approximately
$d_bD_{\mathrm{KV}}$ weights, where
$D_{\mathrm{KV}}=2L H_{\mathrm{KV}}d_h$. Its size therefore follows the
target's layers and KV heads, including the savings from grouped-query
attention, rather than total LLM parameters alone. Sharing head parameters
across layer groups or factorizing the output projection could reduce
this cost; both would change the output space and require validation.
The present measurements establish feasibility through the evaluated 14B
target.}

\begin{table}[htbp]
\centering
\caption{Trainable parameter counts by component (millions). Fusion combines token embeddings with cache features; reinjection supplies cache features to each repair block. Width is $d_b$.}
\label{tab:repair-parameter-groups}
\scriptsize
\setlength{\tabcolsep}{4pt}
{\begin{tabular}{lrrrrrrrr}
\toprule
Target LLM & Width & Encoder & Fusion & Reinjection & Backbone & Head & Other & Total \\
\midrule
Qwen2.5-3B & 256 & 0.44 & 0.66 & 0.33 & 3.29 & 4.74 & 0.00 & 9.46 \\
Qwen2.5-3B & 512 & 1.48 & 1.57 & 1.58 & 15.76 & 9.46 & 0.00 & 29.84 \\
Qwen2.5-3B & 704 & 2.88 & 2.43 & 2.98 & 29.78 & 12.99 & 0.00 & 51.07 \\
Qwen2.5-14B & 320 & 2.76 & 1.84 & 0.51 & 5.14 & 31.56 & 0.00 & 41.81 \\
Qwen2.5-14B & 640 & 9.45 & 4.10 & 2.46 & 24.62 & 63.01 & 0.00 & 103.64 \\
Qwen2.5-14B & 832 & 17.72 & 5.65 & 4.85 & 48.52 & 81.89 & 0.00 & 158.62 \\
\bottomrule
\end{tabular}}
\end{table}

\FloatBarrier

\section{Training Objectives and Functional Effects}
\label{app:training-objectives}
\begin{figure}[t]
\centering
\includegraphics[width=\linewidth]{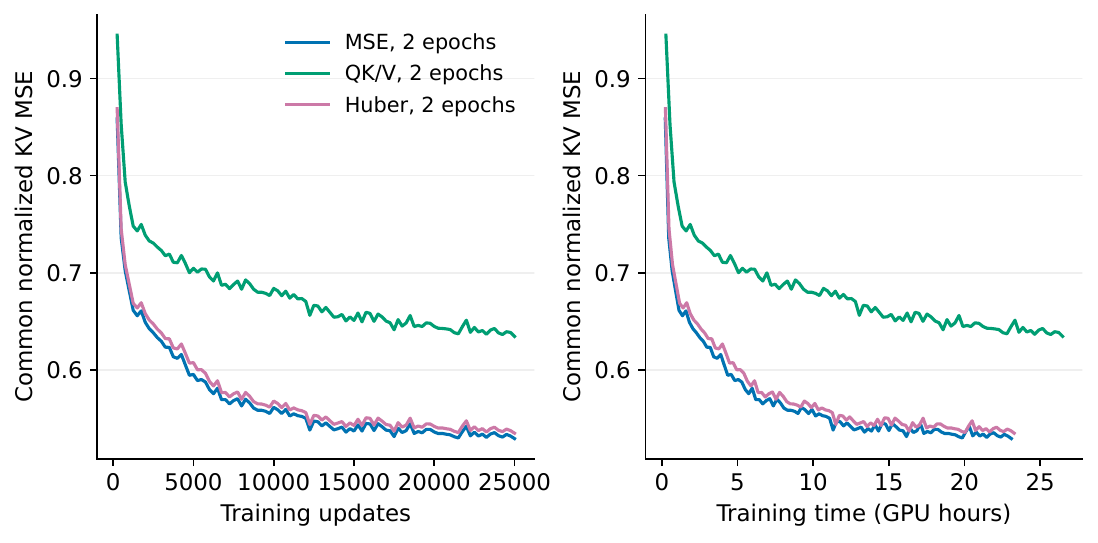}
\caption{Optimization with MSE, QK/V, and Huber on the
same two-epoch training sequence. Both panels report normalized KV MSE
using the shared residual statistics; each point averages 250 updates.
The horizontal axes show updates and measured training time.}
\label{fig:objective-training}
\end{figure}

{
\noindent\textbf{Matched loss comparison.}
Three objectives use the 30M Qwen2.5-3B A0 architecture and the same
50,000 examples, initialization, sample order, normalization statistics,
optimizer, and learning-rate schedule for two epochs (25,000 updates).
Let $e=(\widehat{KV}^{c}-KV^{c}_{\mathrm{joint}})/\sigma_\Delta$ be the
normalized reconstruction error. The objectives are
\begin{align}
\mathcal L_{\mathrm{MSE}} &= \operatorname{mean}(e^2), \\
\mathcal L_{\mathrm{QK/V}} &=
\tfrac12\operatorname{mean}\!\left[
\left(Q_{\mathrm{teacher}}
(\widehat K^{g}-K^{g}_{\mathrm{joint}})^\top/\sqrt{d_h}\right)^2\right]
+\tfrac12\operatorname{mean}(e_V^2), \\
\mathcal L_{\mathrm{Huber}} &= 2\operatorname{mean}(\rho_1(e)),
\end{align}
where $\rho_1$ is the Huber loss with threshold one and superscript $g$
denotes global RoPE coordinates. The teacher queries come from the training
source questions and the frozen target LLM, conditioned on joint document
KV. The QK term averages errors over all query positions, document positions, attention
heads, and target layers. QK/V uses fixed weights of $1/2$ for both terms:
the K term measures attention-logit error in the target LLM's native scale,
whereas the V term uses residual-RMS normalization. These weights specify
the tested objective; they do not equalize the terms' gradient magnitudes.
Each objective uses residual prediction and
block-causal repair attention. Final answer quality is measured on the same
500 MuSiQue requests. All three controlled repairers use BF16 and the same
exact SDPA implementation at each request's actual length. The main
six-epoch checkpoint is an additional serving reference, using the main
evaluation runtime.

MSE has the highest mean F1 in Table~\ref{tab:objective-quality},
exceeding QK/V by 4.78 points and Huber by 1.08 points, based on the
reported mean F1 values; p50 TTFT spans 28.05--28.86\,ms. This controlled comparison
supports MSE as the default among the tested loss formulations and weights.
Calibrating the logit and feature terms offers a further direction for functional objectives.
Figure~\ref{fig:objective-training} compares
optimization using a common normalized KV MSE. All runs use one training
seed; the functional-metric intervals below quantify request-sampling variation.

\noindent\textbf{Functional measurements.}
On the first 256 of these requests, we compare Stale KV, the main six-epoch
repairer, and the three two-epoch repairers. All use the same Full Prefill
continuation, up to 32 tokens, under teacher forcing. We report
$D_{\mathrm{KL}}(p_{\mathrm{full}}\Vert p_{\mathrm{candidate}})$ at the
continuation prediction positions, averaged first within each request.
Attention-output error is measured after the output projection at the last
query position and pooled over layers and requests using squared magnitudes.
Each method receives the same document-token prefix; the trailing document tokens
outside the last complete 16-token cache block and the full query are
processed together.

On these 256 requests, mean KL decreases from 1.731 for Stale KV to 0.555
for the main six-epoch repairer and 0.698 for two-epoch MSE (A0).
Attention-output relative RMSE decreases from 0.598 to 0.328 and 0.353,
respectively. Paired 95\% request-bootstrap intervals for both repairers'
reductions in KL, attention error, and normalized KV MSE lie above zero.
Table~\ref{tab:objective-functional} extends this comparison to QK/V and Huber.
Both also improve all three functional metrics over Stale KV.
Compared with two-epoch MSE, QK/V has higher normalized KV MSE by 0.104
(95\% interval [0.102, 0.107]) and higher KL by 0.218 ([0.124, 0.313]);
Huber's KL difference is $-0.012$ ([$-0.063$, $0.037$]).
Repair brings the target LLM's query processing and predictive
distributions closer to Full Prefill on these requests.}

\begin{table}[htbp]
\centering
\caption{Qwen2.5-3B answer quality on 500 MuSiQue requests with a 30M repairer. The three two-epoch objectives share the same initialization, architecture, examples, sample order, and 25,000 updates. They use exact SDPA at the actual request length. F1 differences are computed from the reported means relative to two-epoch MSE (A0 in Table~\ref{tab:controlled-ablation}). Full, Stale, and the main six-epoch repairer provide serving references from the main evaluation runtime.}
\label{tab:objective-quality}
\scriptsize
\setlength{\tabcolsep}{4pt}
{\begin{tabular}{lrrrrr}
\toprule
Configuration & F1 & EM & $\Delta$F1 (pp) & TTFT (ms, p50) & GPU h \\
\midrule
Full Prefill & 0.3155 & 0.2320 & +6.93 & 68.24 & -- \\
Stale KV & 0.1462 & 0.0720 & -10.00 & 22.41 & -- \\
MSE, 6 epochs & 0.2560 & 0.1700 & +0.98 & 33.28 & -- \\
MSE, 2 epochs & 0.2462 & 0.1720 & +0.00 & 28.86 & 23.10 \\
QK/V, 2 epochs & 0.1984 & 0.1260 & -4.78 & 28.47 & 26.49 \\
Huber, 2 epochs & 0.2354 & 0.1560 & -1.08 & 28.05 & 23.29 \\
\bottomrule
\end{tabular}}
\end{table}

\begin{table}[htbp]
\centering
\caption{Functional effects on the same 256 MuSiQue requests. KL uses teacher forcing along the Full Prefill continuation and is averaged per request. Attention error is measured after the output projection at the last query position and pools squared error across requests and layers. Brackets give 95\% request-bootstrap intervals.}
\label{tab:objective-functional}
\scriptsize
\setlength{\tabcolsep}{4pt}
{\begin{tabular}{lrrr}
\toprule
Configuration & Normalized KV MSE & KL(Full $\Vert$ candidate) & Attention rel. RMSE \\
\midrule
Stale KV & 0.949 [0.930, 0.969] & 1.731 [1.536, 1.936] & 0.598 [0.589, 0.607] \\
MSE, 6 epochs & 0.478 [0.470, 0.488] & 0.555 [0.449, 0.674] & 0.328 [0.315, 0.340] \\
MSE, 2 epochs & 0.502 [0.493, 0.512] & 0.698 [0.576, 0.832] & 0.353 [0.340, 0.367] \\
QK/V, 2 epochs & 0.606 [0.595, 0.618] & 0.916 [0.781, 1.056] & 0.405 [0.390, 0.420] \\
Huber, 2 epochs & 0.508 [0.499, 0.518] & 0.685 [0.569, 0.813] & 0.351 [0.338, 0.365] \\
\bottomrule
\end{tabular}}
\end{table}

\FloatBarrier
\section{Attention Topology and Within-Chunk Context}
\label{app:attention-topology}
{
\noindent\textbf{Controlled training.}
All six reported variants use the same 50,000 training examples, sample order,
shared-parameter initialization, optimizer, and two-epoch budget of 25,000
updates. The effective batch contains four document sequences, with seed 7.
A0--A3 vary only the attention mask: block-causal, strict token-causal,
fully bidirectional, and block-diagonal. A4 supplies stale-cache features
at the backbone entrance; A5 predicts joint KV with a zero output offset.
Token-embedding fusion
is shared by all variants. Removing the per-block conditioning projections
in A4 reduces trainable parameters from 29.8M to 28.3M.

Each variant minimizes squared canonical-KV error with the same frozen
$\sigma_\Delta$ weights. For A5, the normalized target is
$KV^c_{\mathrm{joint}}/\sigma_\Delta$. Training uses FP32 repair
weights and a frozen BF16 target. Evaluation uses BF16 repair and cache
transfer, with exact SDPA attention at the actual request length for all
six variants. We evaluate each final checkpoint on the same 500 MuSiQue
requests. All variants use one training seed.
The large A3 and A5 differences support cross-chunk information exchange
and residual prediction under this budget. A0--A2 and A4 are close in F1:
block-causal attention supplies preceding-chunk context while respecting
document order, and reinjection supplies the original stale features at
each block. These are design rationales; the small observed differences
do not establish a ranking across training seeds.

A5 removes the residual path while holding objective weights and
initialization fixed. Because joint KV and residuals differ in scale,
we additionally evaluate direct prediction with joint-RMS normalization.

\noindent\textbf{Target-matched normalization.}
We train A5-JR to predict $KV^c_{\mathrm{joint}}/\sigma_{\mathrm{joint}}$
with zero output offset. The fixed joint RMS is computed from all
50,000 training examples (99,539,881 document tokens). A5-JR shares
A0/A5's initialization, sample order, optimizer, learning-rate schedule,
and two-epoch budget of 25,000 updates. All final checkpoints are evaluated
on the same 500 MuSiQue requests.
A0, A5, and A5-JR achieve 24.62\%, 0.49\%, and 1.18\% F1,
respectively. The reported means give residual prediction a 23.44-point
advantage over A5-JR; A5-JR improves over A5
by 0.69 points (95\% paired interval [0.13, 1.37], from 10,000
request-bootstrap draws with seed 20260911).
The corresponding unweighted canonical-K/native-V MSE values,
averaged over requests, are 0.109, 0.668, and 0.308,
respectively. Joint-RMS weighting improves this
error for direct prediction, while residual-RMS-weighted MSE increases
from 541.73 to 721.22, reflecting the changed coordinate weights.
A0's residual-RMS-weighted MSE is 0.510.
Residual prediction therefore delivers substantially higher answer F1
than direct prediction under the matched two-epoch training budget.}

{
\noindent\textbf{Attention and regional error.}
A0--A2 have similar boundary and interior error profiles;
block-diagonal attention has higher error in both regions.
Table~\ref{tab:controlled-regions} also separates the first chunk: its K/V
relative RMSE is 0.115/0.157 for A3 and 0.011/0.030 for A0.
A3 still reduces boundary and interior error relative to Stale KV,
but perturbs the first chunk, which includes the chat preamble and has
little missing cross-chunk context. This regional imbalance is a plausible
contributor to its low answer F1: aggregate reconstruction error does not
weight positions by their effect on generation. A0 and A3 use the same
within-chunk attention mask for the first chunk, but learn different weights.
One possible explanation is that A3's shared weights learn local corrections
for later chunks that also perturb the first chunk. This hypothesis is consistent
with the regional errors; the measurements do not isolate it as the cause
of the answer degradation.

\noindent\textbf{Training behavior and cost.}
Under the common normalized KV objective and sample order, the final
250-update mean loss is 0.529 for A0 and 540.9 for A5.
The six reported variants require 23.01--23.31 GPU-hours on one A800;
A5-JR requires 24.30 GPU-hours.}

\noindent\textbf{Applying the predicted K and V.}
Table~\ref{tab:repair-components} uses the fixed six-epoch
30M repairer on the first 256 MuSiQue requests. Applying both predicted
components gives 26.56 F1, compared with 18.81 when only K is repaired and
0.00 when only V is repaired. Each choice combines one jointly trained output with the other
component's stale value, testing output compatibility rather than
separately trained K-only or V-only models. With V-only repair, attention
weights computed from stale K aggregate values produced by the jointly
trained repairer. This changes the pairing of attention weights and value
representations, providing a possible explanation for the loss of answer quality.
Keeping the first chunk's original KV
gives 25.89 F1, a difference of $-0.66$ points with paired interval
[$-2.14$, $0.75$]. Each choice executes the complete repair network.
Thus, keeping the first chunk does not provide an observed quality gain
for this checkpoint, and we retain uniform application of its trained
output. This test uses the standard six-epoch repairer; it does not measure
whether preserving the first chunk would improve A3.
End-to-end latency includes the generated answer; the V-only outputs
reach the 32-token limit on all 256 requests.

\begin{table}[htbp]
\centering
\caption{Output choices for the Qwen2.5-3B 30M repairer on 256 MuSiQue requests. Repair K/V applies the correction to that component; Keep first chunk retains its stale KV. Each variant executes the full repair network. F1 differences are paired against CacheRepair, with 95\% request-bootstrap intervals. Latencies are p50; end-to-end time includes generation of up to 32 tokens.}
\label{tab:repair-components}
\scriptsize
\setlength{\tabcolsep}{4pt}
{\begin{tabular}{lrrrrr}
\toprule
Configuration & F1 (\%) & EM (\%) & $\Delta$F1 (pp) & TTFT (ms) & End-to-end (ms) \\
\midrule
Full Prefill & 33.13 & 25.00 & +6.57 [+2.74, +10.56] & 71.25 & 118.08 \\
Stale KV & 13.52 & 6.64 & -13.03 [-18.35, -7.80] & 23.11 & 77.75 \\
CacheRepair & 26.56 & 17.97 & +0.00 [+0.00, +0.00] & 32.44 & 82.96 \\
Repair K & 18.81 & 14.06 & -7.75 [-12.26, -3.20] & 32.72 & 82.33 \\
Repair V & 0.00 & 0.00 & -26.56 [-31.54, -21.83] & 32.70 & 453.00 \\
Keep first chunk & 25.89 & 18.36 & -0.66 [-2.14, +0.75] & 32.80 & 82.24 \\
\bottomrule
\end{tabular}}
\end{table}

\begin{table}[t]
\centering
\caption{Relative RMSE of K/V for the two-epoch variants on MuSiQue. Boundaries contain the first eight tokens of later chunks; interiors contain the remaining tokens. Errors and reference magnitudes are pooled across requests and layers.}
\label{tab:controlled-regions}
\small
\setlength{\tabcolsep}{4pt}
{
\begin{tabular}{lrrr}
\toprule
Variant & First chunk (K/V) & Boundary (K/V) & Interior (K/V) \\
\midrule
Stale KV & 0.007 / 0.027 & 0.310 / 0.604 & 0.182 / 0.339 \\
A0 CacheRepair & 0.011 / 0.030 & 0.148 / 0.469 & 0.095 / 0.290 \\
A1 Token-causal & 0.017 / 0.037 & 0.150 / 0.470 & 0.095 / 0.290 \\
A2 Bidirectional & 0.021 / 0.038 & 0.149 / 0.469 & 0.095 / 0.290 \\
A3 Block diagonal & 0.115 / 0.157 & 0.191 / 0.503 & 0.126 / 0.310 \\
A4 Entrance-only & 0.010 / 0.029 & 0.149 / 0.468 & 0.095 / 0.289 \\
A5 Direct joint KV & 0.260 / 0.762 & 0.247 / 0.740 & 0.245 / 0.733 \\
\bottomrule
\end{tabular}}
\end{table}

\FloatBarrier

\section{Cross-Model Validation of KV-Error Structure}
\label{app:observation-generality}

We repeat Section~\ref{sec:observations}'s measurements on
Qwen2.5-3B, Llama-3.1-8B, and Qwen2.5-14B using the same first 256 MuSiQue
requests and each target's tokenizer. K is measured in canonical coordinates
and V in native coordinates. We pool squared error and reference magnitude
over the same regions: the first chunk, the first eight tokens of later
chunks, and the remaining later-chunk tokens. Layer--position maps use
16 relative-position bins. Table~\ref{tab:cross-model-regions} gives
request-bootstrap intervals.

\begin{figure}[htbp]
\centering
\includegraphics[width=\linewidth]{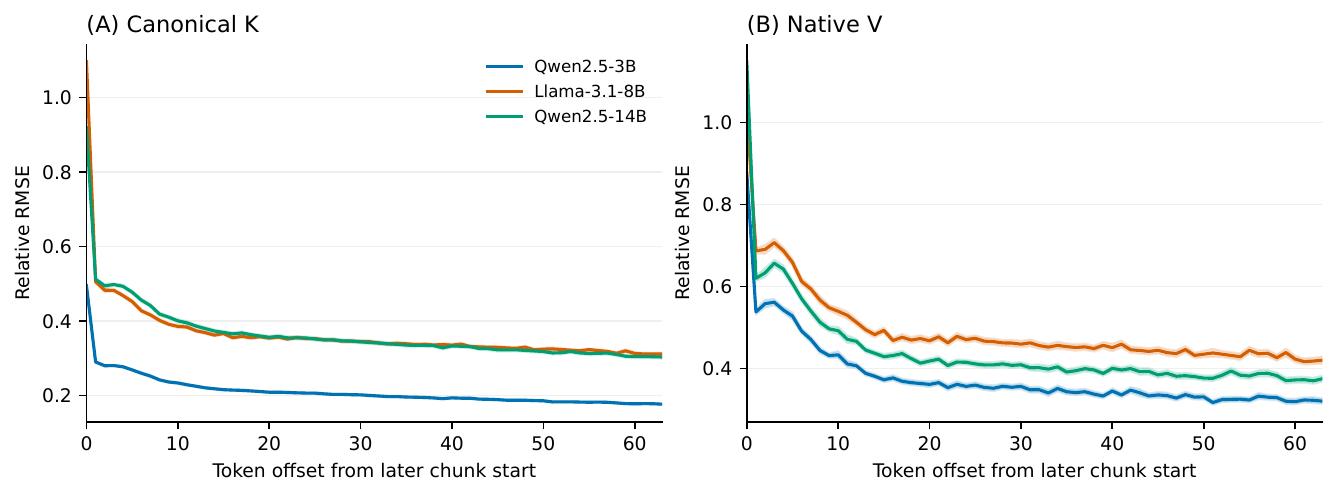}
\caption{Stale KV error by distance from a later chunk's
start, on the same 256 MuSiQue requests. Curves pool layers and heads;
shading gives 95\% request-bootstrap intervals.}
\label{fig:cross-model-boundary}
\end{figure}

\begin{figure}[htbp]
\centering
\includegraphics[width=\linewidth]{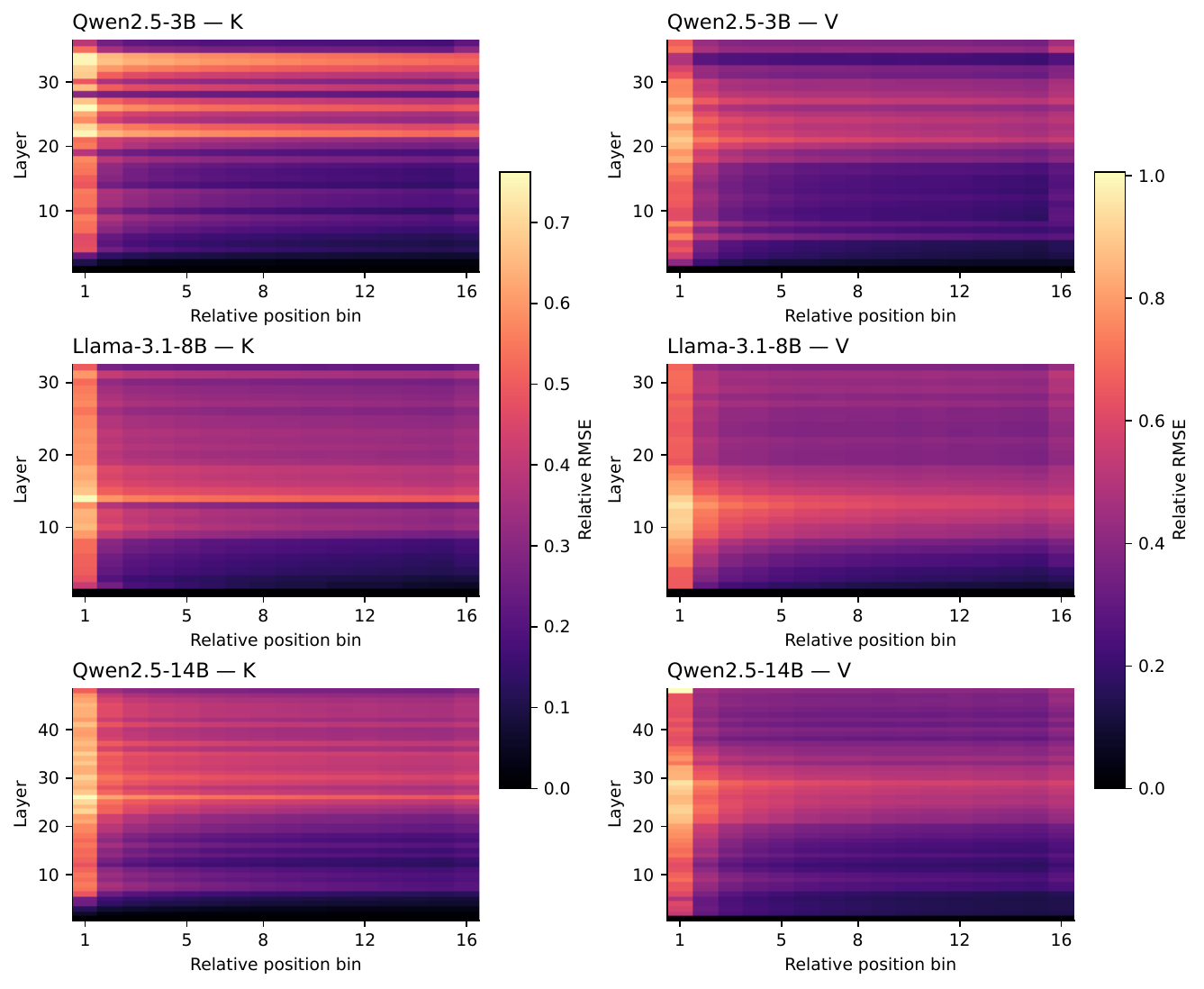}
\caption{Stale KV relative RMSE by target-LLM layer and
position within later chunks, using the same 256 MuSiQue requests.
Columns show canonical K and native V, with shared color scales across
models. The 16 position bins reveal both boundary hotspots and
error bands extending through chunk interiors.}
\label{fig:cross-model-position}
\end{figure}

Figure~\ref{fig:cross-model-position} shows boundary hotspots
and layer-persistent interior errors in all three target LLMs, supporting
repair across document positions. Boundary relative RMSE is
1.71--1.79$\times$ the interior value for K and 1.67--1.83$\times$ for V
(Table~\ref{tab:cross-model-regions}). Figure~\ref{fig:cross-model-boundary}
shows how error decays with distance from the chunk boundary.

\begin{table}[htbp]
\centering
\caption{Stale-cache relative RMSE on the same 256 MuSiQue requests. Boundary and interior refer to later chunks; the boundary is the first eight tokens. Brackets give 95\% request-bootstrap intervals.}
\label{tab:cross-model-regions}
\scriptsize
\setlength{\tabcolsep}{4pt}
{\begin{tabular}{llrrr}
\toprule
Target LLM & KV & First chunk & Boundary & Interior \\
\midrule
Qwen2.5-3B & K & 0.007 [0.007, 0.007] & 0.309 [0.307, 0.310] & 0.181 [0.179, 0.183] \\
Qwen2.5-3B & V & 0.027 [0.026, 0.027] & 0.603 [0.600, 0.607] & 0.336 [0.332, 0.340] \\
Llama-3.1-8B & K & 0.011 [0.011, 0.011] & 0.572 [0.570, 0.574] & 0.320 [0.317, 0.322] \\
Llama-3.1-8B & V & 0.026 [0.025, 0.026] & 0.728 [0.724, 0.732] & 0.437 [0.432, 0.442] \\
Qwen2.5-14B & K & 0.011 [0.010, 0.011] & 0.552 [0.549, 0.554] & 0.311 [0.306, 0.315] \\
Qwen2.5-14B & V & 0.026 [0.025, 0.027] & 0.713 [0.709, 0.718] & 0.391 [0.385, 0.396] \\
\bottomrule
\end{tabular}}
\end{table}

\FloatBarrier

\section{Learned-Fusion Baseline Compatibility}
\label{app:learned-baseline-compatibility}

\noindent\textbf{Target-matched KV Packet checkpoints.}
Each KV Packet wrapper contains eight header and eight trailer
embeddings and is evaluated with its corresponding frozen target LLM.
All three target LLMs use the same 512 examples from CacheRepair's generic
corpus, 30 epochs, effective batch size 64, 240 updates, and seed 42.
The learning rate decays linearly from $5\times10^{-4}$ to zero.
Each target uses its native chat template and supplies teacher continuations
and logits for distillation. The calibration size and schedule follow
KV Packet's 256--512-example, 30-epoch recipe~\citep{chen2026kvpacket}.
We compare downstream quality and TTFT under each method's training recipe,
using a shared source corpus for cross-dataset transfer.

\noindent\textbf{Cached-prefix execution.}
We prefill the chat preamble once and each wrapped document
chunk independently. We place the preamble cache before the wrapped chunks,
concatenate the physical KV entries in request order, and apply the target's global positional
encoding when loading the cache for the query. Learned header and trailer
entries remain in the physical cache and contribute to online transfer and
query-attention costs. Independent chunk compilation and pinned-host preload
are offline; online TTFT includes transfer, positional encoding, paged-cache
injection, query processing, and first-token generation. All target LLMs
remain frozen during downstream evaluation.

\noindent\textbf{Calibration budget.}
For Qwen2.5-3B, we expand KV Packet's calibration set
from 512 to 2,048 and 8,192 nested examples from the same generic corpus.
Each run uses 30 epochs, batch size 64, and seed 42, giving 240, 960,
and 3,840 updates. Thus both data and optimization budgets increase.
Evaluation uses the same four datasets and implementation
(Table~\ref{tab:packet-calibration-size}). The 512-to-8192 changes span
$-0.04$ to $+2.84$ F1 points; all 95\% intervals from 10,000 paired
request-bootstrap draws include zero.
CacheRepair 30M retains higher mean F1 on MuSiQue, HotpotQA, and TriviaQA;
KV Packet retains higher mean F1 on MultiHop-RAG. The main Pareto curves
use the 512-example configuration.

\begin{table}[H]
\centering
\caption{KV Packet calibration budget on Qwen2.5-3B. F1 (\%) uses the same 500 requests per dataset. $\Delta$ is 8192 minus 512, in percentage points, with pointwise 95\% paired-bootstrap intervals. CR is the fixed 30M CacheRepair reference.}
\label{tab:packet-calibration-size}
\small
\setlength{\tabcolsep}{4pt}
{\begin{tabular}{lrrrrr}
\toprule
Dataset & 512 & 2048 & 8192 & $\Delta$ [95\% CI] & CR \\
\midrule
MuSiQue & 19.05 & 19.79 & 21.90 & +2.84 [-0.03, +5.69] & 25.60 \\
HotpotQA & 39.42 & 41.35 & 40.89 & +1.47 [-1.61, +4.42] & 53.92 \\
MultiHop-RAG & 55.09 & 54.53 & 55.05 & -0.04 [-2.50, +2.45] & 53.58 \\
TriviaQA & 72.26 & 72.83 & 73.58 & +1.32 [-1.27, +3.88] & 78.26 \\
\bottomrule
\end{tabular}}
\end{table}

\section{Scaling and Amortization Details}
\label{app:scaling-amortization}

\noindent\textbf{Document length and system cost.}
We evaluate all 18 configurations at 1K, 2K, 4K, 8K,
and 16K document tokens for each target LLM, keeping 16 chunks.
CacheBlend and InfoFlow use fixed recomputation fractions; EPIC uses fixed
per-chunk token budgets. The 16 source-passage streams and queries are
fixed, with each length taking a prefix under the target's tokenizer and
including its chat preamble. After warmup, we time three runs per request
with identical context capacity and KV pools across methods. Bootstrap
resampling preserves each request's three runs; crossover brackets use
adjacent measured lengths. These workloads measure first-token cost.

\begin{figure}[htbp]
\centering
\includegraphics[width=\linewidth]{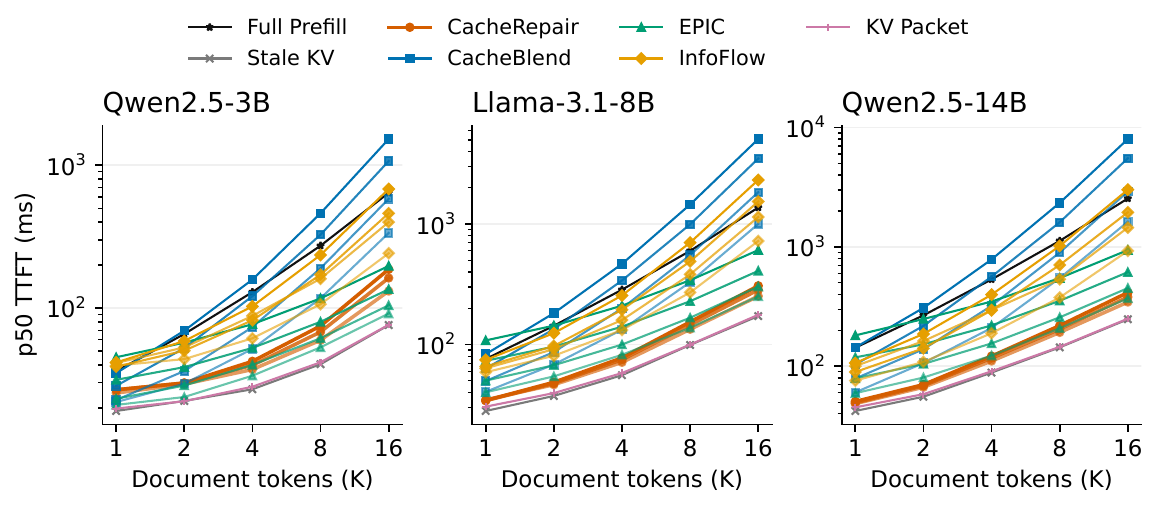}
\caption{p50 TTFT across document lengths for the three
target LLMs, with 16 chunks and all 18 main-evaluation configurations.
Each point contains three measurements of each of 16 fixed requests.
Darker curves within a method indicate larger repairers or higher
recomputation budgets. Both axes are logarithmic.}
\label{fig:length-ttft}
\end{figure}

\begin{figure}[htbp]
\centering
\includegraphics[width=\linewidth]{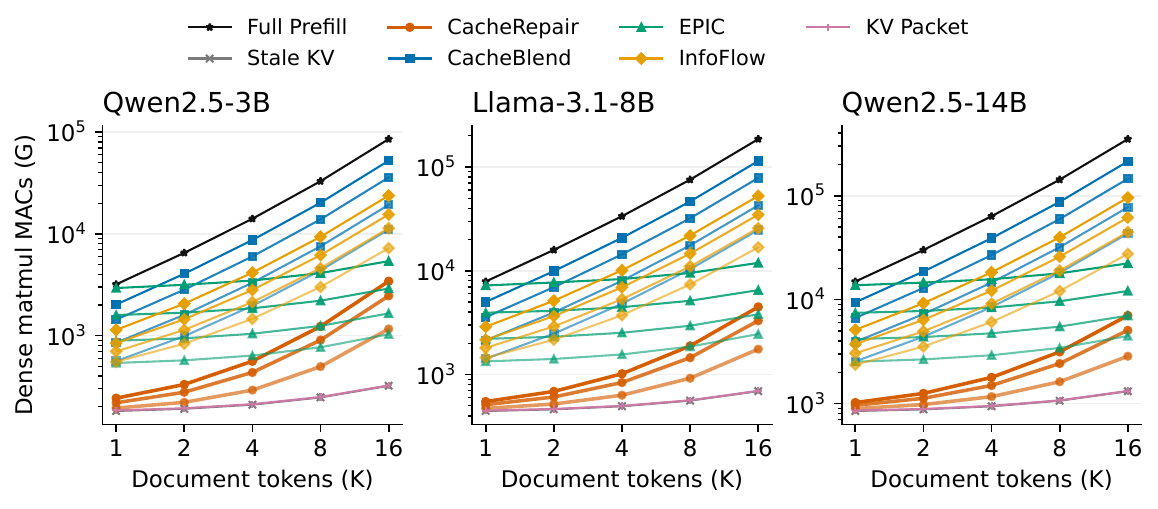}
\caption{Dense matrix MACs across document lengths,
for the same configurations as Figure~\ref{fig:length-ttft}.
Counts include target-LLM passes, repair, scoring, and first-token output
heads. G denotes $10^9$ MACs; both axes are logarithmic.}
\label{fig:length-macs}
\end{figure}

\begin{figure}[htbp]
\centering
\includegraphics[width=\linewidth]{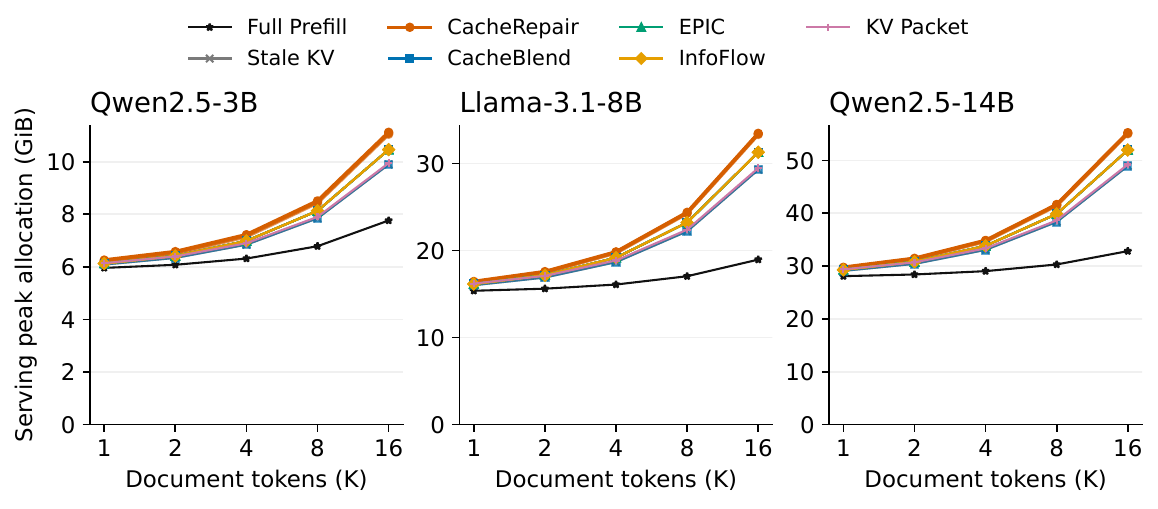}
\caption{Peak serving-worker memory across document
lengths, including target-LLM weights, KV pools, and online temporary
tensors. Each point is the maximum across 48 timed runs; each worker
loads one repairer capacity. Styles follow Figure~\ref{fig:length-ttft}.}
\label{fig:length-serving-memory}
\end{figure}

Figures~\ref{fig:length-ttft}, \ref{fig:length-macs}, and
\ref{fig:length-serving-memory} show latency, computation, and memory.
Dense MACs count executed matrix operations, including selected-token
passes, scoring, and repair padding. TTFT additionally includes transfers,
elementwise operations, normalization, and cache writes. Peak memory
includes target-LLM weights, the KV pool, and temporary tensors, with one
repairer loaded per worker. Additional allocation is relative to the same
target LLM's idle Full Prefill worker. For Qwen2.5-14B at 8K and 16K,
caches are prepared before serving and the offline compiler stays on CPU.
Memory counters report framework allocations.
For Qwen2.5-14B at 16K tokens, CacheRepair 159M peaks at 55.25\,GiB,
compared with 48.95\,GiB for Stale KV, a difference of 6.30\,GiB.
The reported 24.32\,GiB increment is measured against the idle Full Prefill
worker and includes cache and runtime allocations as well as the repairer;
it is not the repair network's parameter footprint.

Across 270 configurations and 12,960 timed runs,
88 of 108 repairer--selective-baseline pairs have lower repair p50 TTFT
at all five lengths. At 16K, the largest repairers achieve
3.43--6.14$\times$ speedups over Full Prefill using 2.0--4.0\% of its
dense MACs; Table~\ref{tab:length-resources} provides the exact resource
measurements. Qwen2.5-3B's 30M and 51M repairers become faster than
CacheBlend~0.10 between 2K and 4K, supported by paired intervals at both
endpoints. Conversely, EPIC8 becomes faster than Qwen2.5-14B's 159M repairer
between 4K and 8K: its budget remains 120 document tokens, whereas repair
processes every token.
This comparison separates coverage from cost: a fixed fraction increases
the number of recomputed tokens with length, whereas a fixed boundary
budget covers a decreasing fraction. The crossover measures their
runtime tradeoff; the first-token study does not attach answer quality
to those operating points.

\begin{table}[htbp]
\centering
\caption{System cost at 16K document tokens and 16 chunks (16 requests, three measurements each). We show the largest repairers and the fixed baseline presets used in the chunk experiment. TMAC denotes $10^{12}$ dense matrix MACs. Peak is the serving-worker allocation, including weights, KV pool, and temporary tensors; Extra is its increment above the paired idle Full Prefill worker. Both memory columns are in GiB.}
\label{tab:length-resources}
\small
\setlength{\tabcolsep}{4pt}
{\begin{tabular}{llrrrr}
\toprule
Target LLM & Configuration & TTFT p50 (ms) & TMACs & Peak & Extra \\
\midrule
Qwen2.5-3B & Full Prefill & 640.58 & 85.51 & 7.76 & 1.26 \\
Qwen2.5-3B & Stale KV & 76.06 & 0.32 & 9.90 & 3.39 \\
Qwen2.5-3B & CacheRepair 51M & 186.82 & 3.42 & 11.12 & 4.62 \\
Qwen2.5-3B & CacheBlend 0.1 & 335.45 & 11.05 & 9.90 & 3.39 \\
Qwen2.5-3B & EPIC 16 & 104.46 & 1.65 & 10.46 & 3.96 \\
Qwen2.5-3B & InfoFlow 0.1 & 399.21 & 11.37 & 10.46 & 3.96 \\
Qwen2.5-3B & KV Packet 8+8 & 76.35 & 0.32 & 9.95 & 3.45 \\
Llama-3.1-8B & Full Prefill & 1373.07 & 185.68 & 18.95 & 1.69 \\
Llama-3.1-8B & Stale KV & 172.29 & 0.69 & 29.27 & 12.02 \\
Llama-3.1-8B & CacheRepair 93M & 307.22 & 4.48 & 33.45 & 16.20 \\
Llama-3.1-8B & CacheBlend 0.1 & 1003.97 & 24.76 & 29.27 & 12.02 \\
Llama-3.1-8B & EPIC 16 & 302.68 & 3.81 & 31.28 & 14.03 \\
Llama-3.1-8B & InfoFlow 0.1 & 1145.57 & 25.75 & 31.28 & 14.03 \\
Llama-3.1-8B & KV Packet 8+8 & 176.50 & 0.70 & 29.46 & 12.21 \\
Qwen2.5-14B & Full Prefill & 2546.12 & 350.22 & 32.83 & 1.90 \\
Qwen2.5-14B & Stale KV & 247.71 & 1.32 & 48.95 & 18.02 \\
Qwen2.5-14B & CacheRepair 159M & 414.83 & 7.04 & 55.25 & 24.32 \\
Qwen2.5-14B & CacheBlend 0.1 & 1627.95 & 43.28 & 48.95 & 18.02 \\
Qwen2.5-14B & EPIC 16 & 451.86 & 7.02 & 51.96 & 21.03 \\
Qwen2.5-14B & InfoFlow 0.1 & 1452.79 & 44.62 & 51.96 & 21.03 \\
Qwen2.5-14B & KV Packet 8+8 & 250.74 & 1.32 & 49.23 & 18.30 \\
\bottomrule
\end{tabular}}
\end{table}

\FloatBarrier

\noindent\textbf{Answer quality with longer contexts.}
\phantomsection\label{app:long-quality}
We test length extrapolation with the fixed six-epoch Qwen2.5-3B 30M
repairer, whose training sequences reach 2,269 tokens
(Appendix~\ref{app:generic-corpus}). On 128 MultiHop-RAG requests,
CacheRepair reaches 46.68/49.02/49.80\% F1 at 4K/8K/16K, with
3.36/4.08/4.45$\times$ speedups over Full Prefill.
At 16K, CacheRepair and CacheBlend 0.6 have close mean F1 (49.80\% and
49.41\%), while CacheRepair reduces p50 TTFT from 1,487.4 to 145.8\,ms
($10.20\times$). These results measure the checkpoint's quality--latency
tradeoff at sequence lengths beyond its training range.
Table~\ref{tab:long-quality-latency} reports all seven configurations,
including KV Packet's faster operating points with higher mean F1 at
4K/8K and lower mean F1 at 16K. Paired intervals relative to Full Prefill
use 10,000 request-bootstrap draws with seed 20260911.
Appendix~\ref{app:additional-requests} explains how answer content
and wording affect F1 ordering, including cases where cache reuse scores
above Full Prefill.

We construct nested contexts from query-only
BM25 retrieval (top 64 passages, at most two per article). We take the
first 128 eligible requests in the fixed 500-request order after applying
Appendix~\ref{app:cohort-construction}'s rules to the expanded inputs.
All lengths use the same questions; text segments contain at most 256
tokens, with the chat preamble added to the first segment.
The resulting contexts contain 16--19, 32--36, and 64--70 chunks,
respectively. Each of the seven configurations uses identical inputs at
each length and the original greedy, 32-token decoding protocol, yielding
2,688 generations. These measurements extend both context and chunk count;
the separate system-cost sweep above holds chunk count at 16.

Canonicalizing target keys removes their local positional rotation;
the repair network's own positional encoding still operates over the
assembled sequence. Extending the training distribution to longer
retrieved contexts is a natural next step for learning residual corrections
over these larger position ranges.

\begin{table}[htbp]
\centering
\caption{Quality and latency on the same 128 MultiHop-RAG requests with Qwen2.5-3B and nested retrieved contexts. All seven evaluated configurations are shown. TTFT is in ms (p50); F1 is in percent. The final row gives CacheRepair minus Full Prefill F1 with pointwise 95\% paired request-bootstrap intervals at each length.}
\label{tab:long-quality-latency}
\small
\setlength{\tabcolsep}{5pt}
\begin{tabular}{lrrrrrr}
\toprule
& \multicolumn{2}{c}{4K} & \multicolumn{2}{c}{8K} & \multicolumn{2}{c}{16K} \\
\cmidrule(lr){2-3}\cmidrule(lr){4-5}\cmidrule(lr){6-7}
Method & F1 & TTFT & F1 & TTFT & F1 & TTFT \\
\midrule
Full Prefill & 48.63 & 129.9 & 50.59 & 274.8 & 52.54 & 648.1 \\
Stale KV & 50.13 & 27.6 & 42.29 & 43.2 & 37.19 & 74.8 \\
CacheRepair 30M & 46.68 & 38.7 & 49.02 & 67.4 & 49.80 & 145.8 \\
KV Packet 512 & 52.15 & 26.2 & 50.78 & 41.4 & 48.05 & 75.7 \\
CacheBlend 0.6 & 49.41 & 161.1 & 50.59 & 456.3 & 49.41 & 1487.4 \\
EPIC 8 & 46.35 & 35.0 & 46.95 & 65.0 & 44.52 & 148.1 \\
EPIC 64 & 49.02 & 79.9 & 50.31 & 204.1 & 47.15 & 621.3 \\
\midrule
$\Delta$F1 (95\% CI) & \multicolumn{2}{c}{\scriptsize -1.95 [-7.03, +3.12]} & \multicolumn{2}{c}{\scriptsize -1.56 [-5.47, +2.34]} & \multicolumn{2}{c}{\scriptsize -2.73 [-8.98, +3.52]} \\
\bottomrule
\end{tabular}
\end{table}

\noindent\textbf{Cache payload.}
A BF16 cache with $N$ document tokens contains
$2ND_{\mathrm{KV}}$ bytes, where $D_{\mathrm{KV}}$ includes both K and V
across target layers and KV heads. At 4,096 tokens, the payloads are
144, 512, and 768\,MiB for Qwen2.5-3B, Llama-3.1-8B, and Qwen2.5-14B.
KV Packet additionally stores its learned boundary-token entries.
Online measurements load pinned-host caches onto one GPU; the reported
H2D times therefore characterize the host-to-device path on this A800
node. Transfer time depends on effective host-to-device bandwidth
and cache residency; Table~\ref{tab:repair-latency-components} separates
this cost from repair execution. Framework buffers and concurrent offline preparation are accounted
for separately in the memory measurements.

\noindent\textbf{Chunk count.}
On MuSiQue500, we keep every document token, its order, and
the query fixed, and partition the document stream into 4, 8, 12, or 16
contiguous chunks of approximately equal length. The complete chat preamble
stays in the first chunk. We compare Stale KV, CacheRepair 30M at epoch six,
CacheBlend 0.10, EPIC16, InfoFlow 0.10, and KV Packet 8+8. Full Prefill uses
the identical token stream and provides a shared reference across counts.
All metrics use these same 500 requests.

\begin{figure}[htbp]
\centering
\includegraphics[width=\linewidth]{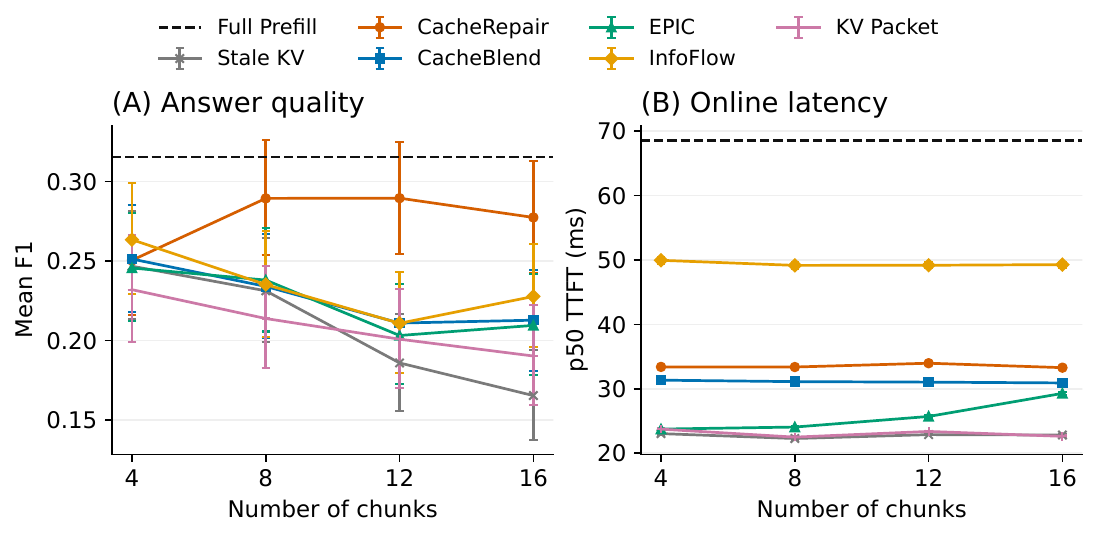}
\caption{Chunk-count sensitivity for Qwen2.5-3B on the
same 500 MuSiQue requests. Document tokens, order, and queries remain
fixed. CacheRepair uses the 30M epoch-six checkpoint; baseline settings are specified above. Error bars are pointwise
95\% request-bootstrap intervals. The dashed line is the shared Full
Prefill reference.}
\label{fig:chunk-count-sensitivity}
\end{figure}

Figure~\ref{fig:chunk-count-sensitivity} shows the quality and latency trends.
At 8, 12, and 16 chunks, CacheRepair gains 5.81, 10.36, and 11.20 F1
percentage points over Stale KV; paired 95\% bootstrap intervals are
[2.35, 9.35], [6.74, 13.94], and [7.67, 14.80], respectively.
At four chunks, the gain is 0.37 points with an interval of [$-2.95$, 3.72].
CacheRepair's p50 TTFT ranges from 33.29 to 33.97\,ms across the four
counts, compared with 68.54\,ms for Full Prefill. Repair benefits increase with chunk count at fixed content.

\FloatBarrier

\noindent\textbf{Chunk order.}
Context position can affect answer quality \citep{liu2024lostmiddle}. We reorder the same document chunks on 256 MuSiQue requests, using the fixed 30M checkpoint. Reversing
the chunks gives 25.47 F1 and a fixed permutation gives 26.49, compared
with 26.56 in retrieval order. Full Prefill also responds to ordering,
with F1 ranging from 28.82 to 33.13. Relative to the corresponding Full
Prefill output, the change in CacheRepair's F1 difference is $+3.22$
points for reversal (95\% paired interval [$-2.14$, $8.54$]) and $+0.57$
for permutation ([$-4.99$, $6.16$]). Both intervals include zero for these
two predetermined orderings.

\noindent\textbf{Concurrent requests.}
\phantomsection\label{app:concurrent-workload}
With Qwen2.5-3B and the 30M repairer, CacheRepair throughput increases
from 11.35 to 30.70 requests/s as concurrency rises from one to eight,
compared with 7.45 to 11.35 for Full Prefill. At concurrency eight,
CacheRepair serves $2.70\times$ as many requests per second as Full
Prefill; KV Packet reaches 46.80 requests/s.
Table~\ref{tab:concurrent-workload} pairs throughput with p95 first-token
latency for all four methods.

We use one vLLM engine on an A800 with the first 64 requests from each
downstream dataset, giving a fixed 256-request workload. A closed-loop
client keeps 1, 2, 4, or 8 requests outstanding; each completion admits
the next request. The target LLM batches active requests, while the 30M
six-epoch repairer processes document caches per request within scheduler
batches. Cache-based methods use warm pinned-host caches; all methods
use the same 32-token generation cap and identical KV-pool and engine limits.
Table~\ref{tab:concurrent-workload} measures completed requests per second
over the workload, including fill and drain, and submit-to-first-token
latency. It complements the single-request timing in the main evaluation.

The measured implementation motivates further integration of repair with
the serving scheduler. Future system optimizations include batching
variable-length repair inputs with independent
request masks, and pipelining H2D transfer, repair, and KV writes across
requests. Such a pipeline must preserve each request's transfer-before-repair
dependency while overlapping independent work. Jointly scheduling these
stages is a direction for improving the throughput and tail latency of
concurrent cache fusion.

\begin{table}[htbp]
\centering
\caption{Closed-loop serving on 256 requests (64 per downstream dataset), Qwen2.5-3B, one A800. Each cell reports throughput (requests/s) and p95 submit-to-first-token latency (ms). $C$ is the number of outstanding requests. One complete workload run is measured per method and concurrency level.}
\label{tab:concurrent-workload}
\small
\setlength{\tabcolsep}{7pt}
\begin{tabular}{lrrrr}
\toprule
Method & $C=1$ & $C=2$ & $C=4$ & $C=8$ \\
\midrule
Full Prefill & 7.45 / 130.6 & 8.96 / 219.3 & 10.42 / 312.5 & 11.35 / 521.9 \\
Stale KV & 12.43 / 29.5 & 19.48 / 50.3 & 31.61 / 65.7 & 44.61 / 96.7 \\
CacheRepair 30M & 11.35 / 52.2 & 16.90 / 74.2 & 24.25 / 105.4 & 30.70 / 173.9 \\
KV Packet 512 & 13.16 / 27.5 & 19.71 / 47.2 & 32.69 / 59.0 & 46.80 / 85.0 \\
\bottomrule
\end{tabular}
\end{table}

\FloatBarrier

\noindent\textbf{One-time training cost.}
The nine epoch-six repairers require 65.9--130.5 A800
GPU-hours each, measured from training logs through update 75,000.
These totals include repeated work after a checkpoint restart and in-loop
checkpointing. Statistics construction takes 8.92/11.56/19.11 GPU-hours
for Qwen2.5-3B/Llama-3.1-8B/Qwen2.5-14B and is shared across the three
capacities of each target. FP32 deployment files occupy 36.2--605.9 MiB,
including normalization buffers and metadata.

\begin{table}[htbp]
\centering
\caption{One-time costs of the nine epoch-six repairers. Training time covers recorded launches through the epoch-six checkpoint. Statistics are built once per target LLM and shared across its capacities. Checkpoint sizes are FP32 deployment files. The last column gives a latency-equivalent reuse count, using mean TTFT savings over the equally weighted four downstream datasets.}
\label{tab:training-amortization}
\small
\setlength{\tabcolsep}{4pt}
{\begin{tabular}{llrrrrr}
\toprule
Target LLM & Repairer & Training & Statistics & File & Mean saving & Reuse \\
 & (M params.) & (GPU h) & (GPU h) & (MiB) & (ms/request) & (M requests) \\
\midrule
Qwen2.5-3B & 9 & 65.9 & 8.92 & 36.2 & 53.3 & 5.06 \\
Qwen2.5-3B & 30 & 70.1 & 8.92 & 114.0 & 50.4 & 5.65 \\
Qwen2.5-3B & 51 & 69.4 & 8.92 & 195.0 & 48.6 & 5.80 \\
Llama-3.1-8B & 23 & 81.9 & 11.56 & 89.1 & 126.9 & 2.65 \\
Llama-3.1-8B & 59 & 84.5 & 11.56 & 225.0 & 124.9 & 2.77 \\
Llama-3.1-8B & 93 & 86.9 & 11.56 & 355.6 & 123.7 & 2.86 \\
Qwen2.5-14B & 42 & 129.3 & 19.11 & 160.3 & 265.0 & 2.02 \\
Qwen2.5-14B & 104 & 129.8 & 19.11 & 396.1 & 263.9 & 2.03 \\
Qwen2.5-14B & 159 & 130.5 & 19.11 & 605.9 & 260.5 & 2.07 \\
\bottomrule
\end{tabular}}
\end{table}

For one deployed capacity, let $H_{\mathrm{train}}$ and
$H_{\mathrm{stats}}$ denote training and shared statistics GPU-hours.
Using mean TTFTs in seconds, equally weighted across the four datasets,
the latency-equivalent reuse count is
{
\begin{equation}
\widehat R =
\frac{3600(H_{\mathrm{train}}+H_{\mathrm{stats}})}
{\overline{T}_{\mathrm{full}}-\overline{T}_{\mathrm{repair}}}.
\label{eq:training-amortization}
\end{equation}}
This proxy assumes a serial stream on one A800 and counts
first-token latency savings. Cached-document preparation, decoding,
batching, and GPU utilization determine deployment-specific amortization.

\section{Generic Repair Pretraining Corpus Composition}
\label{app:generic-corpus}

We construct the generic repair pretraining corpus from six
sources using the fixed query and passage resources released with
CoRAG \citep{wang2025corag}, built on KILT \citep{petroni2021kilt}: ELI5, FEVER,
Natural Questions (NQ), Wizard of Wikipedia (WoW), T-REx, and Structured
Zeroshot. Queries and retrieval IDs come from \texttt{corag/kilt};
passage texts come from \texttt{corag/kilt-corpus}.\footnote{\url{https://huggingface.co/datasets/corag/kilt} and \url{https://huggingface.co/datasets/corag/kilt-corpus}.}
Each example pairs a source query with its published first-$k$ passages
($k\in[2,20]$), preserving text and retrieval order. The corpus contains
50,000 examples, 874,000 chunks, and 99,539,881 document tokens under the
Qwen2.5-3B-Instruct tokenizer. Downstream selection follows
Appendix~\ref{app:cohort-construction}.

Document length is the total number of retrieved-passage tokens, excluding
queries, separators, and special tokens. Table~\ref{tab:generic-corpus-length}
reports both the sampling bands and the actual token ranges. Within every
band, ELI5, FEVER, NQ, and WoW each contribute 20\% of examples;
T-REx and Structured Zeroshot each contribute 10\%.

\begin{table}[H]
    \centering

    \caption{Training examples by document length. Counts exclude query tokens; shares use 50,000 examples and 99,539,881 document tokens.}
    \label{tab:generic-corpus-length}
    \small
    \setlength{\tabcolsep}{4.5pt}
    \begin{tabular}{lrrrrrr}
        \toprule
        Category & Band definition & Realized range & Rows & Row share & Document tokens (share) & Mean/row \\
        \midrule
        Short  & 1--1,024     & 261--1,013   & 5,000  & 10.0\% &  3,532,712 (3.55\%) &   706.5 \\
        Medium & 1,025--2,048 & 1,376--1,812 & 5,000  & 10.0\% &  8,080,516 (8.12\%) & 1,616.1 \\
        Long   & 2,049--4,096 & 2,131--2,269 & 40,000 & 80.0\% & 87,926,653 (88.33\%) & 2,198.2 \\
        \midrule
        Total  & 1--4,096     & 261--2,269   & 50,000 & 100.0\% & 99,539,881 (100.00\%) & 1,990.8 \\
        \bottomrule
    \end{tabular}
\end{table}

\end{document}